%% file: main.tex
\documentclass[10pt,twocolumn,letterpaper]{article}

\usepackage[algorithms]{wacv}      

\input{preamble}
\usepackage[accsupp]{axessibility}  

\definecolor{wacvblue}{rgb}{0.21,0.49,0.74} \usepackage[pagebackref,breaklinks,colorlinks,allcolors=wacvblue]{hyperref}

\def\wacvPaperID{2436} 
\def\confName{WACV}
\def\confYear{2026}

\title{Controllable Long-term Motion Generation with Extended Joint Targets}

\author{
Eunjong Lee \qquad
Eunhee Kim \qquad
Sanghoon Hong \qquad
Eunho Jung \qquad
Jihoon Kim\thanks{Corresponding author.} \\
Cinamon Inc.\\
Seoul, South Korea\\
{\tt\small
\{eunjong, eunhee, sanghoon, eunho, jihoon\}@cinamon.io}
}

\begin{document}
\maketitle

\input{sec/0_abstract}
\input{sec/1_intro}
\input{sec/2_related_work}
\input{sec/3_method}
\input{sec/4_experiments}

\input{sec/5_conclusion}
{
    \small
    \bibliographystyle{ieeenat_fullname}
    \bibliography{main}
}

\end{document}


\maketitle

\renewcommand{\thesection}{\Alph{section}}

\section{Dataset Details}

COMET is trained on two complementary motion-capture datasets: AMASS and CIRCLE. AMASS provides diverse, long-horizon behaviors essential for modeling natural navigation, while CIRCLE offers densely annotated, contact-rich reaching motions. Both datasets are converted into a unified SMPL-X representation.

\subsection{AMASS}

The AMASS dataset~\cite{AMASS:ICCV:2019} serves as the primary source for learning general human motion patterns. It is a large-scale repository containing over 17,000 sequences spanning a broad range of movements, with extensive coverage of locomotion behaviors. This diversity is crucial for training COMET to synthesize natural navigation dynamics, particularly when the character must traverse large distances to reach a target. Since AMASS lacks explicit goal annotations, we introduce a \emph{pseudo-goal} strategy by randomly selecting a future frame and using its ground-truth joint position as the target. This enables COMET to acquire goal-directed behaviors from unannotated sequences. To ensure quality and consistency, we remove clips with excessive foot lifting and uniformly resample all motions to 30 frames per second (fps).

\subsection{CIRCLE}

The CIRCLE dataset~\cite{araujo2023circle} provides task-specific data for learning fine-grained goal-reaching behaviors. It contains approximately 10 hours of motion capture data, comprising more than 7,000 sequences from five subjects performing whole-body reaching tasks with both the right and left hands. The motions include a wide variety of reaching-related actions, such as bending, crawling, crouching, and kneeling, performed within a static environment. Unlike AMASS, CIRCLE offers explicit annotations of initial and target conditions, which are critical for teaching COMET precise upper-body and arm movements necessary for accurate interaction with target locations.

\subsection{Preprocessing}

All motion clips undergo a shared preprocessing pipeline. First, we compute joint positions for translation-dependent features and to re-ground each sequence so that the lowest joint touches $z{=}0$. Second, sequences are trimmed into 8 second windows that match the temporal receptive field of the transformer modules. Third, we compute dataset-wide statistics for every feature channel (root motion, joint rotations, goal descriptors, and joint coordinates), which are later used to standardize the inputs at training and inference time. Finally, the combined motion dataset is randomly partitioned into disjoint splits with an 80/10/10 ratio for training, validation, and testing. This pipeline guarantees that heterogeneous sources provide consistent supervision for COMET's goal-aware motion synthesis.

\section{Implementation Details}
\subsection{Model Architecture}

The COMET architecture is built upon a , where both the encoder and decoder are realized using Transformer Encoder modules. The latent space is configured with a dimensionality of $ z = 64 $. Input motion features, including pose and delta ($\delta$) representations, are first processed through a 16-layer Multi-Layer Perceptron (MLP). Simultaneously, the conditional intention vector, which provides task-specific guidance, is projected via a single linear layer before integration into the main Transformer pipeline. 
The Transformer modules employ standard sinusoidal positional encoding and are configured as detailed in Table~\ref{tab:comet-architecture}.

\begin{table}[tbp]
\centering
\small
\setlength{\tabcolsep}{8pt}
\caption{Key architectural specifications of COMET.}
\label{tab:comet-architecture}
\begin{tabular}{lc}
\toprule
\textbf{Component} & \textbf{Specification} \\
\midrule
Latent space dimensionality ($z$) & 64 \\
MLP layers (pose + delta features) & 16 \\
Regularization & LayerNorm + Dropout \\
Conditional intention embedding & Linear layer (1 layer) \\
Transformer layers & 4 \\
Model dimension & 64 \\
Attention heads & 8 \\
Feed-forward network dimension & 64 \\
Positional encoding & Sinusoidal \\
\bottomrule
\end{tabular}
\end{table}

This design enables COMET to effectively model complex, temporally coherent motion sequences while maintaining computational efficiency suitable for real-time generation.

\subsection{Training Details}
The COMET model was trained using the PyTorch on a workstation equipped with four NVIDIA A6000 GPUs.  The full training process required approximately 36 hours to complete a total of 1,800 epochs. 
We used a batch size of 512 and maintained a fixed learning rate of $1\times10^{-4}$ throughout training. 
To mitigate exposure bias, which commonly arises in autoregressive sequence generation, we employed a \textbf{scheduled sampling} strategy. This strategy was activated at epoch $10$, after which the number of autoregressive (AR) steps was progressively increased over subsequent epochs until epoch $50$, where it was capped at a maximum of $10$ AR steps. This gradual scheduling allowed the model to smoothly transition from relying on ground-truth inputs to generating longer sequences based on its own predictions, thereby improving stability and long-horizon motion synthesis.

\section{Reference-guided Feedback Details}

\begin{algorithm}[tb]
\caption{Reference-Guided Feedback (RGF)}
\label{alg:reference-guided-feedback}
\small
\begin{algorithmic}[1]
\Require COMET decoder $p_{\theta}$, initial pose $\mathbf{p}_1$, control joints $J_c$,
\Statex \hspace{\algorithmicindent} joint goals $\{\mathbf{G}_j\}_{j \in J_c}$, GMM params $\{(\mu_k,\Sigma_k)\}_{k=1}^{K}$,
\Statex \hspace{\algorithmicindent} feedback scalar $\alpha$, sequence length $T$, stop distance $d_{\text{goal}}$
\State Initialize sequence $\hat{\mathcal{M}} \leftarrow \{\mathbf{p}_1\}$, \quad $\text{feedback\_active} \leftarrow \text{True}$
\vspace{2pt}

\For{$i = 1, \dots, T-1$}
    \State Update intentions $\mathbf{I}_{i}$
    \State Predict delta:
    \Statex \hspace{\algorithmicindent} $\hat{\mathbf{\delta}}_{i+1} \leftarrow p_{\theta}(\mathbf{\delta}_{i+1}|\mathbf{p}_i,\mathbf{I}_i,\mathbf{z}_i \sim \mathcal{N}(\mathbf{0},\mathbf{I}))$
    \State Pose update: $\hat{\mathbf{p}}_{i+1} \leftarrow \mathbf{p}_i + \hat{\mathbf{\delta}}_{i+1}$
    \vspace{4pt}

    \If{$\text{feedback\_active}$}
        \State Extract features: $\hat{\mathbf{f}}_{i+1} \subset \hat{\mathbf{p}}_{i+1}$
        \State Select closest GMM component:
        \[
           k^{*} = \arg\min_{k} \sqrt{(\hat{\mathbf{f}}_{i+1}-\mu_k)^\mathrm{T} 
           \Sigma_k^{-1} (\hat{\mathbf{f}}_{i+1}-\mu_k)}
        \]

        \State \textbf{Correct features:} 
        \[
            \hat{\mathbf{f}}_{i+1}^{\text{corrected}} = 
            \hat{\mathbf{f}}_{i+1} + \alpha (\mu_{k^*} - \hat{\mathbf{f}}_{i+1})
        \]

        \State \textbf{Update pose with corrected features:}
        \[
            \hat{\mathbf{p}}_{i+1} = \text{Update}\big(\hat{\mathbf{p}}_{i+1}, \hat{\mathbf{f}}_{i+1}^{\text{corrected}}\big)
        \]

        \vspace{3pt}
        \If{$\|\mathbf{P}_{i+1}^{xy}-\mathbf{G}_{\text{avg}}^{xy}\|_2 < d_{\text{goal}}$}
            \State $\text{feedback\_active} \leftarrow \text{False}$
        \EndIf
    \EndIf
    \State $\hat{\mathcal{M}} \leftarrow \hat{\mathcal{M}} \cup \{\hat{\mathbf{p}}_{i+1}\}$
\EndFor
\vspace{3pt}
\State \Return generated sequence $\hat{\mathcal{M}}$
\end{algorithmic}
\end{algorithm}

To model the distribution of plausible human poses, we train a Gaussian Mixture Model (GMM) using walking motion data from the training set. The GMM is configured with 50 mixture components and optimized using the Expectation-Maximization (EM) algorithm, with a maximum of 1,000 iterations to ensure convergence. For our benchmark model, training typically completes in under one minute.

Algorithm~\ref{alg:reference-guided-feedback} outlines the full Reference-Guided Feedback (RGF) process. At each timestep, COMET predicts a delta pose that updates the current state. When feedback is active, the predicted joint rotations and pelvis height are softly corrected toward the closest GMM component mean, as determined by the Mahalanobis distance. This correction step (lines 10--12) nudges the motion toward the learned manifold of natural human poses, mitigating drift and preserving long-term stability. The feedback loop is deactivated once the character enters the vicinity of the goal, allowing for precise final adjustments without interference from the reference signal.

\subsection{Choice of GMM Components}
To determine the appropriate number of Gaussian components \(K\) for the Reference-Guided Feedback (RGF) module, we conducted an experiment by varying \(K\) over a wide range. As shown in Table~\ref{tab:gmm-component-ablation}, performance saturates beyond \(K = 50\) and slightly declines at \(K = 100\). Therefore, we select \(K = 50\) as the setting for our benchmark model.

\begin{table}[tbp]
\centering
\small
\setlength{\tabcolsep}{2pt} 
\caption{Ablation on the number of Gaussian components $K$ for the Reference-Guided Feedback (RGF).}
\label{tab:gmm-component-ablation}
\begin{tabular*}{\columnwidth}{@{\extracolsep{\fill}}lccc@{}}
\toprule
\# Components $K$ & SR (\%) ($\uparrow$) & FS (\%) ($\downarrow$) & DTG (cm) ($\downarrow$) \\
\midrule
0   & 31.47 & 18.97 & 59.25 \\
3   & 49.20 & 13.62 & 21.00 \\
5   & 51.36 & 13.63 & 21.23 \\
10  & 51.25 & 13.93 & 21.44 \\
30  & 49.44 & \textbf{13.59} & 23.16 \\
\textbf{50}  & \textbf{52.67} & 13.95 & \textbf{20.30} \\
100 & 51.73 & 14.03 & 21.68 \\
\bottomrule
\end{tabular*}
\end{table}

\subsection{Feedback Scale $\alpha$}

The feedback scale parameter \(\alpha\) is a critical hyperparameter in COMET, as it controls the strength of the reference-guided feedback applied to the generated motion. The value of \(\alpha\) directly influences performance. When \(\alpha\) is too small, the corrective signal becomes weak, leading to insufficient error correction and reduced motion stability. In this case, the generated poses may drift away from the learned distribution of plausible human motions. Conversely, an excessively large \(\alpha\) can cause the model to overfit to specific reference poses, resulting in mode collapse and degraded motion quality, such as exaggerated foot skating artifacts. Through empirical evaluation, we found that \(\alpha = 0.01\) provides the best balance between stability and naturalness, delivering consistent improvements across all benchmarks. For the motion in-betweening task, we use a higher value of \(\alpha = 0.05\), where the reference pose is fixed to the given target pose to ensure precise convergence to the final keyframe.

\subsection{Feedback Stopping Distance}

In COMET, the reference-guided feedback is applied during the main transit phases of motion but is intentionally deactivated as the model approaches its target. This design prevents interference with the final fine-grained adjustments needed to accurately align the controlled joints with their target positions. Persisting with feedback near the goal can be counterproductive, as it may push the motion toward the closest reference pose rather than the precise target configuration. We define the feedback stopping distance as the threshold at which feedback is disabled. Empirically, we determined that a threshold of 1 meter yields the optimal trade-off, ensuring both accurate goal attainment and natural, stable final postures.

\begin{figure*}[tbp]
\centering
\setlength{\tabcolsep}{4pt}

\begin{subfigure}[b]{0.95\linewidth}
    \centering
    \small
    \caption{Success Rate (SR, \%) ($\uparrow$)}
    \label{tab:multi_joint_sr}
    \begin{tabular*}{\linewidth}{@{\extracolsep{\fill}}lccccccc@{}}
    \toprule
    \# Joints & Pelvis & L. Ankle & R. Ankle & Head & L. Wrist & R. Wrist & Mean \\
    \midrule
    1 & 94.80 & 66.93 & 64.13 & 86.53 & 66.13 & 51.20 & 71.62 \\
    2 & 94.06 & 65.39 & 54.42 & 74.70 & 72.39 & 61.53 & 70.41 \\
    3 & 96.38 & 68.98 & 52.64 & 69.78 & 72.45 & 66.73 & 71.16 \\
    4 & 94.80 & 66.49 & 51.29 & 64.03 & 72.82 & 67.07 & 69.42 \\
    5 & 95.11 & 62.80 & 49.36 & 66.86 & 73.42 & 64.94 & 68.75 \\
    6 & 95.33 & 59.87 & 49.47 & 69.60 & 73.07 & 65.87 & 68.87 \\
    \bottomrule
    \end{tabular*}
\end{subfigure}

\vspace{0.5em}

\begin{subfigure}[b]{0.95\linewidth}
    \centering
    \small
    \caption{Foot Skate (FS) ($\downarrow$)}
    \label{tab:multi_joint_fs}
    \begin{tabular*}{\linewidth}{@{\extracolsep{\fill}}lccccccc@{}}
    \toprule
    \# Joints & Pelvis & L. Ankle & R. Ankle & Head & L. Wrist & R. Wrist & Mean \\
    \midrule
    1 & 15.94 & 13.74 & 14.32 & 13.80 & 12.07 & 12.33 & 13.70 \\
    2 & 15.27 & 15.05 & 15.24 & 14.82 & 14.73 & 14.52 & 14.94 \\
    3 & 15.26 & 15.29 & 15.23 & 14.89 & 14.75 & 14.84 & 15.04 \\
    4 & 15.29 & 15.49 & 15.36 & 15.41 & 15.27 & 15.32 & 15.36 \\
    5 & 15.46 & 15.54 & 15.47 & 15.50 & 15.44 & 15.49 & 15.48 \\
    6 & 15.32 & 15.32 & 15.32 & 15.32 & 15.32 & 15.32 & 15.32 \\
    \bottomrule
    \end{tabular*}
\end{subfigure}

\vspace{0.5em}

\begin{subfigure}[b]{0.95\linewidth}
    \centering
    \small
    \caption{Distance-to-Goal (DTG, cm) ($\downarrow$)}
    \label{tab:multi_joint_dtg}
    \begin{tabular*}{\linewidth}{@{\extracolsep{\fill}}lccccccc@{}}
    \toprule
    \# Joints & Pelvis & L. Ankle & R. Ankle & Head & L. Wrist & R. Wrist & Mean \\
    \midrule
    1 & 3.30 & 12.56 & 12.84 & 8.81 & 26.41 & 34.02 & 16.32 \\
    2 & 4.16 & 11.94 & 16.89 & 9.60 & 11.30 & 16.15 & 11.67 \\
    3 & 3.02 & 10.08 & 15.42 & 9.40 & 9.11 & 11.83 & 9.81 \\
    4 & 3.45 & 9.98 & 14.14 & 10.34 & 9.19 & 11.61 & 9.79 \\
    5 & 3.16 & 10.17 & 13.60 & 9.47 & 8.63 & 11.37 & 9.40 \\
    6 & 2.67 & 10.57 & 13.36 & 8.17 & 8.33 & 11.37 & 9.08 \\
    \bottomrule
    \end{tabular*}
\end{subfigure}

\caption{Evaluation results for multi-joint control with varying numbers of simultaneously controlled joints.}
\label{fig:multi_joint_control}
\end{figure*}

\section{Evaluation Details}
\label{appendix:evaluation-details}

\subsection{Single-Joint Control (Right Wrist)}

The single-joint control experiment assesses the model's ability to generate motion from unseen initial poses towards specific target goals. This setup follows the evaluation methodology employed in WANDR \cite{diomataris2024wandr}, utilizing goals uniformly distributed within a 3D cylindrical space. This approach allows for an effective validation of the model's generalization capabilities to inputs not encountered during training. The specific experimental parameters are as follows: (1) Angle, 360 degrees divided into 5 uniform increments (i.e., 72-degree intervals). (2) Height, the range from 0.5m to 1.8m, divided into 5 uniform increments. (3) Distance, the range from 0.5m to 5m, divided into 5 uniform increments. (4) Initial Pose: 6 distinct unseen initial poses are used. (5) For each combination of the above parameters, 5 trials are sampled. The total number of cases is $5 \times 5 \times 5 \times 6 \times 5 = 3750$. The duration for all generated motions is set to 8 seconds. In this setup, the orientation intention is defined as the direction from the current pelvis position towards the target goal.

\subsection{Multi-Joint Control}
To evaluate COMET's ability to control multiple joints simultaneously, we sampled random target poses from the held-out test set.  
Target positions for the pelvis in the $XY$ plane were selected from five discrete directions and five distance settings, following the same protocol as in the single-joint control evaluation. The vertical ($Z$) position was implicitly determined by the selected final pose.  
We used six unseen final target poses, with each paired with six different initial poses. For each unique target--initial pairing, five trials were conducted, providing a comprehensive evaluation across diverse joint configurations.  
Metrics were first computed individually for each controlled joint and then averaged to report the final performance. Each motion sequence was fixed to 8 seconds in duration.

\subsection{Long-term Sequential Target Control}
To evaluate the capacity for sustained long-distance locomotion and its adaptability to dynamically shifting objectives, we introduce a method for evaluation, navigating towards a sequence of three consecutive target goals, each positioned at a fixed right wrist height of 1.0 meter. Each segment of the sequence requires covering a distance of 5.0 meters. The evaluation is conducted using 6 distinct initial poses, randomly selected from the test set. For each initial pose, the motion generation involves selecting one of 5 uniform directions for each of the three consecutive target segments. This methodology yields $5^3 = 125$ unique locomotion paths per initial pose, resulting in a total of $6 \times 125 = 750$ distinct test cases. For each test case, 5 trials are sampled, bringing the total number of evaluated motions to $750 \times 5 = 3750$. The duration for each of the three motion segments is set to 8 seconds.

\subsection{Evaluation for Motion In-betweening}
For the motion in-betweening task, we follow the evaluation protocol introduced by Harvey et al.~\cite{harvey2020robust}, with several key modifications. While the original study evaluated models only on fixed-length intervals, our evaluation considers in-betweening over arbitrary intervals up to the maximum sequence length supported by the models. In each trial, the first and last keyframes of the interval are provided to the models as boundary conditions. For CondMDI~\cite{cohan2024condmdi}, we ensured fairness by training it exclusively on scenarios where the first and last frames are given, rather than allowing it to learn arbitrary frame predictions. For DNO~\cite{karunratanakul2024optimizing}, we adopted the ODE step size of 100 as reported in the original paper and generated outputs using both 100-step and 300-step optimization settings for comparison. Both CondMDI~\cite{cohan2024condmdi} and DNO~\cite{karunratanakul2024optimizing} were reimplemented and retrained under identical conditions using the same training dataset as COMET. Furthermore, their generation window was set to 240 frames (8 seconds) to match the temporal receptive field used during COMET’s training.

\subsection{Evaluation for Plug-and-Play Motion Stylization}
As stylization is difficult to quantify using conventional numeric metrics, classifier-based measures such as Style Recognition Accuracy (SRA) first introduced in Motion Puzzle~\cite{jang2022motion} have been also adopted in MoST~\cite{kim2024most} and SMooDi~\cite{zhong2024smoodi}. However, these classifier-based metrics are fundamentally misaligned with the design of COMET, limiting their reliability as evaluators. Instead, stylization was assessed qualitatively through a user-preference study conducted on domain experts, including motion-capture specialists and professional animators.

The reliability of classifier-based metrics in motion stylization, namely SRA \cite{jang2022motion}, acquires reliability based on two assumptions:
(1) training and evaluation use labeled data sets that share a common classification scheme, ensuring results are compared under the same classification criterion; 
(2) because both the model and the metric depend on the same scheme, even if one trains on a labeled data set different from the one used in the evaluation, the model tends to achieve similar scores, providing reproducibility. Motion Puzzle~\cite{jang2022motion}, MoST \cite{kim2024most} and SMooDi~\cite{zhong2024smoodi} follow this setting and, in that context, appropriately report quantitative results with SRA. In contrast, COMET’s Reference-Guided Feedback does not rely on a pre-defined label taxonomy. It rapidly learns style motions composed of the required actions and injects them in a plug-and-play manner, reducing dependence on large labeled datasets and improving practical usability. Moreover, in motion data the conceptual separation between content and style is often inherently ambiguous or mutually entangled, so the chosen classification scheme can strongly influence classifier-based metrics. Consequently, such metrics are valid for comparing models that share the same scheme but are not suitable for evaluating COMET, which operates independently of dataset-specific style taxonomies. Additionally, since COMET operates without a content-labeling scheme, it is incompatible with metrics that presuppose content classification, such as Style Consistency (SC)~\cite{wen2021autoregressive}. Therefore, stylization evaluation relies on a user study that qualitatively compares the results obtained with the same style input motion to SMooDi.

\begin{figure}[tbp]
  \centering

  \begin{subfigure}{0.48\linewidth}
    \centering
    \includegraphics[width=\linewidth]{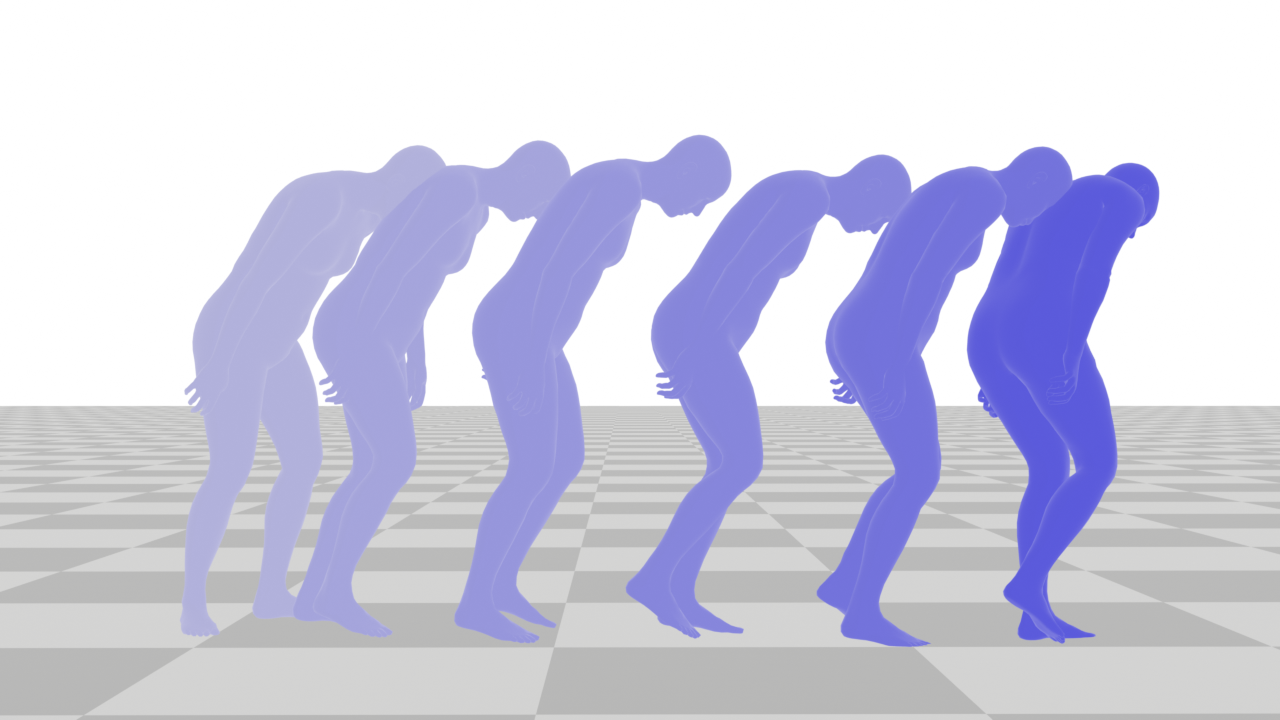} 
    \caption{100STYLE BentForward}
    \label{fig:sup-100style-bentforward}
  \end{subfigure}\hfill
  \begin{subfigure}{0.48\linewidth}
    \centering
    \includegraphics[width=\linewidth]{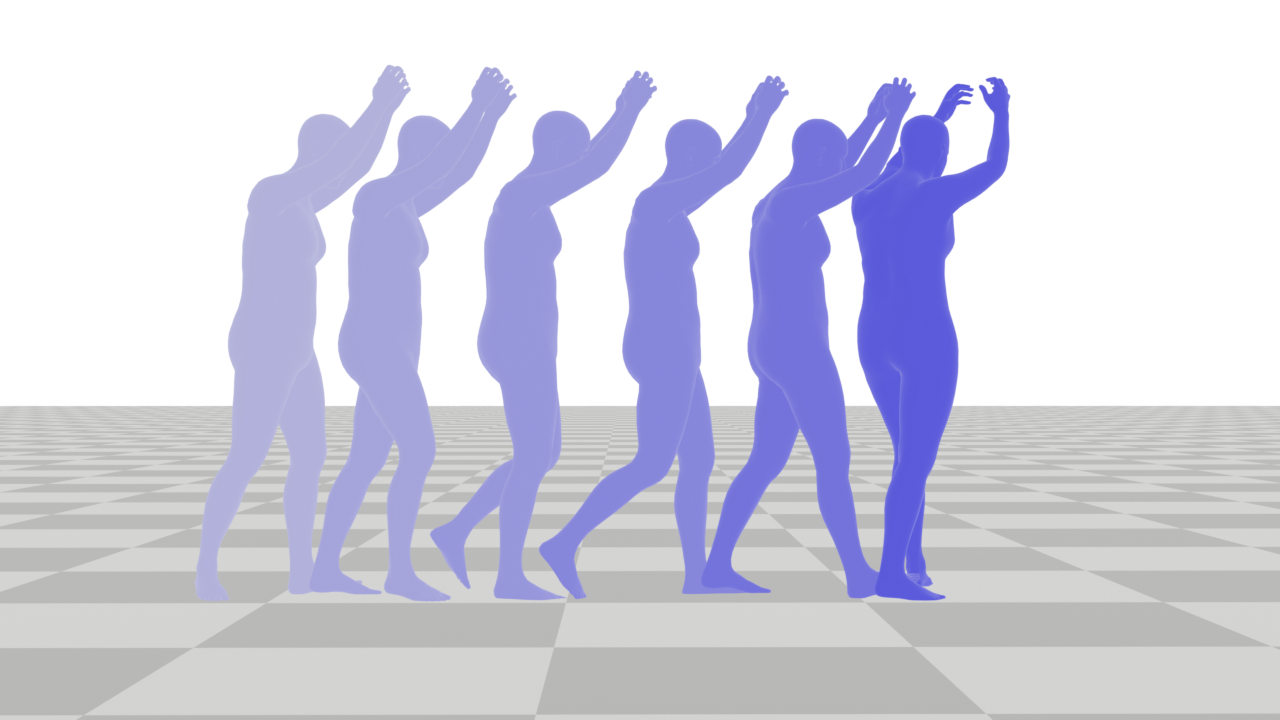}
    \caption{100STYLE ArmsAboveHead}
    \label{fig:sup-100style-armsabovehead}
  \end{subfigure}

  \vspace{0.6em}

  \begin{subfigure}{\linewidth}
    \centering
    \includegraphics[width=\linewidth]{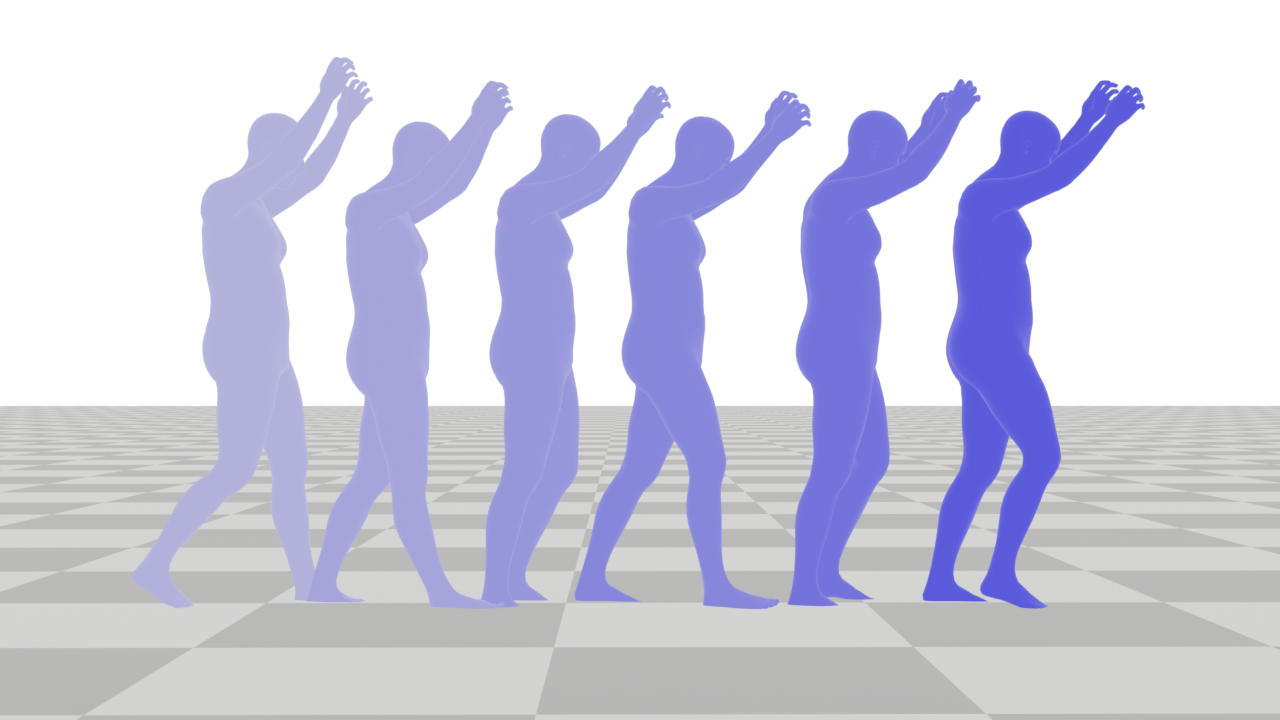}
    \caption{COMET generated motion with ArmsAoveHead as style input}
    \label{fig:sup-comet-armsabovehead}
  \end{subfigure}

  \caption{Comparison between Original 100STYLE dataset and stylized motion generated with COMET. Despite COMET faithfully capturing the style, a classifier with a one-layer Transformer architecture similar with used in SMooDi \cite{zhong2024smoodi}, classified COMET’s output illustrated in \subref{fig:sup-comet-armsabovehead} as BentForward. This suggests that the classifier can easily overfit to non-salient cues rather than the primary characteristics.
}
  \label{fig:three-panel}
\end{figure}

\section{User Preference Study}

While we make extensive efforts to quantitatively evaluate motion quality using established metrics, these measures may not fully capture perceptual differences as judged by humans. To address this gap, we conducted a user preference study to directly assess the perceptual quality of our generated motions. A total of 30 participants with professional experience in motion-related fields, including motion capture engineers, 3D artists, and character animators, were recruited to ensure the reliability of the subjective evaluation. The study consisted of two tasks:

\paragraph{Motion In-betweening.} 
Participants were presented with side-by-side comparisons of motion sequences generated by three methods: COMET, CondMDI~\cite{cohan2024condmdi}, and DNO~\cite{karunratanakul2024optimizing}. All methods were evaluated under identical boundary conditions to ensure fairness. Each participant viewed 20 test cases and was asked to select the motion they perceived as having the highest overall quality. To reduce noise from uncertain judgments, a \emph{``Not sure''} option was provided for cases where the differences between sequences were ambiguous.

\paragraph{Motion Stylization.} 
For stylization, COMET was compared against the baseline method SMoodi~\cite{zhong2024smoodi}. Participants evaluated 24 generated sequences, corresponding to 12 distinct target styles with two sequences generated per style. Each pair consisted of one motion produced by COMET and one by the baseline, and participants were asked to choose which output exhibited higher stylistic quality and naturalness. Similar to the in-betweening task, a \emph{``Not sure''} option was included to handle ambiguous cases where no clear preference could be determined.

\begin{figure}
    \centering
    \includegraphics[width=\linewidth]{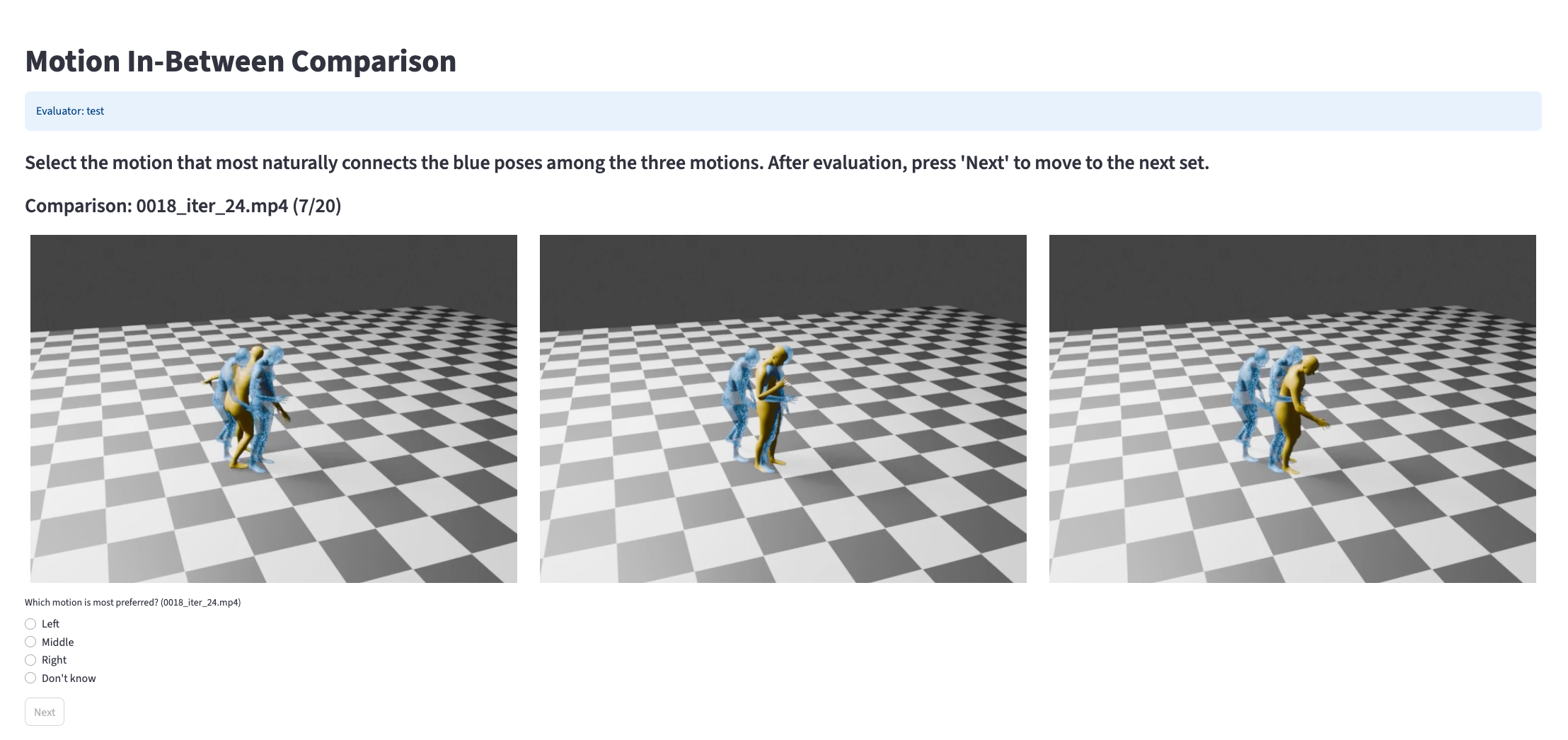}
    \caption{Motion In-betweening User Study}
    \label{fig:sup-comet-mib-userstudy}
\end{figure}

\begin{figure}
    \centering
    \includegraphics[width=\linewidth]{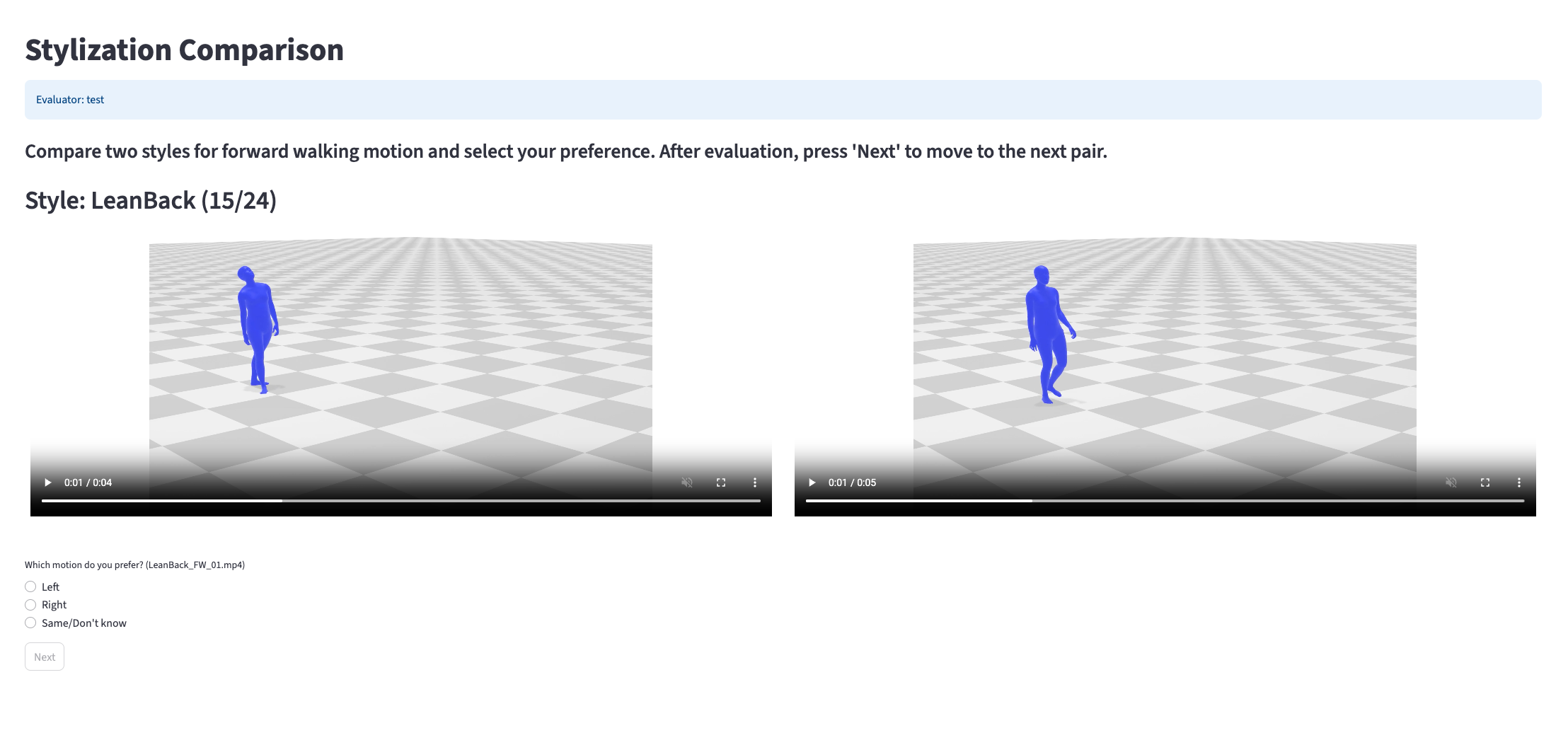}
    \caption{Stylization User Study}
    \label{fig:sup-comet-style-userstudy}
\end{figure}

\section{Comparison with Diffusion-based Joint Control Model}

\begin{table}[tbp]
\centering
\small
\setlength{\tabcolsep}{4pt} 
\caption{Motion generation speed comparison (seconds). Lower values indicate faster generation. 
For OmniControl, the time remains nearly constant as it generates the entire sequence in a single pass, 
regardless of sequence length within its processing capacity. 
DartControl results are measured under its optimization-based setting.}
\label{tab:speed_comparison}
\begin{tabular*}{\columnwidth}{@{\extracolsep{\fill}}lccc@{}}
\toprule
Method & 30 Frames & 60 Frames & 150 Frames \\
\midrule
WANDR        & $0.51 \pm 0.07$   & $1.00 \pm 0.14$   & $2.42 \pm 0.18$ \\
OmniControl  & $57.13 \pm 2.82$  & $62.49 \pm 4.56$  & $56.76 \pm 1.77$ \\
DartControl  & $82.04 \pm 0.23$  & $160.44 \pm 1.22$ & $397.81 \pm 1.22$ \\
Ours (COMET) & $0.64 \pm 0.18$   & $1.30 \pm 0.19$   & $3.20 \pm 0.23$ \\
\bottomrule
\end{tabular*}
\end{table}

We adopt a Variational Autoencoder (VAE) as the core architecture of COMET due to its superior efficiency, which enables real-time controllable motion generation. In contrast, recent diffusion-based approaches, while achieving remarkable accuracy and generating fine-grained motion details, incur substantially higher sampling costs. For example, OmniControl~\cite{xie2024omnicontrol} requires 1,000 denoising steps to generate a sequence, making it roughly 90 times slower than COMET for producing 30 frames. DartControl~\cite{zhao2024dartcontrol} reduces the denoising process to 10 DDIM~\cite{song2020denoising} steps but introduces an additional 100-step iterative optimization procedure to refine motion quality. This repetitive optimization causes the sampling time to grow linearly with sequence length. Consequently, diffusion-based methods are powerful but are better suited for offline generation scenarios, where their significant computational demands are less restrictive. A detailed speed comparison is provided in Table~\ref{tab:speed_comparison}.

{
    \small
    \bibliographystyle{ieeenat_fullname}
    \bibliography{main}
}

%% file: preamble.tex
\usepackage{algorithm}
\usepackage{amsmath}
\usepackage{float}
\usepackage{algpseudocode}
\usepackage{multirow}
\usepackage{makecell}
\usepackage{booktabs}
\usepackage{amssymb}
\newcommand{\cmark}{\checkmark}

%% file: sec/0_abstract.tex
\begin{abstract}
Generating stable and controllable character motion in real-time is a key challenge in computer animation. Existing methods often fail to provide fine-grained control or suffer from motion degradation over long sequences, limiting their use in interactive applications. We propose COMET, an autoregressive framework that runs in real time, enabling versatile character control and robust long-horizon synthesis. Our efficient Transformer-based conditional VAE allows for precise, interactive control over arbitrary user-specified joints for tasks like goal-reaching and in-betweening from a single model. To ensure long-term temporal stability, we introduce a novel reference-guided feedback mechanism that prevents error accumulation. This mechanism also serves as a plug-and-play stylization module, enabling real-time style transfer. Extensive evaluations demonstrate that COMET robustly generates high-quality motion at real-time speeds, significantly outperforming state-of-the-art approaches in complex motion control tasks and confirming its readiness for demanding interactive applications. Video results and code are available at~\url{https://comet-proj.github.io/}.
\end{abstract}

%% file: sec/1_intro.tex
\section{Introduction}
\label{sec:1_introduction}

The generation of realistic human motion under precise user control is a foundational problem in computer graphics and animation. For interactive applications such as virtual reality, games, and robotics, the ideal system must not only synthesize natural-looking movements but also respond to high-level control signals in real time. A critical aspect of this control involves satisfying specific spatial constraints, such as directing an avatar’s hand to a target or guiding the avatar’s body into a sitting posture. The primary difficulty lies in balancing natural motion synthesis with strict adherence to spatial targets. This challenge is magnified as the complexity of control specifications increases, making it more difficult to satisfy these constraints under the stringent demands of real-time applications.

\begin{figure}[t]
\centering

\begin{subfigure}[b]{0.48\linewidth}
    \centering
    \includegraphics[width=\linewidth]{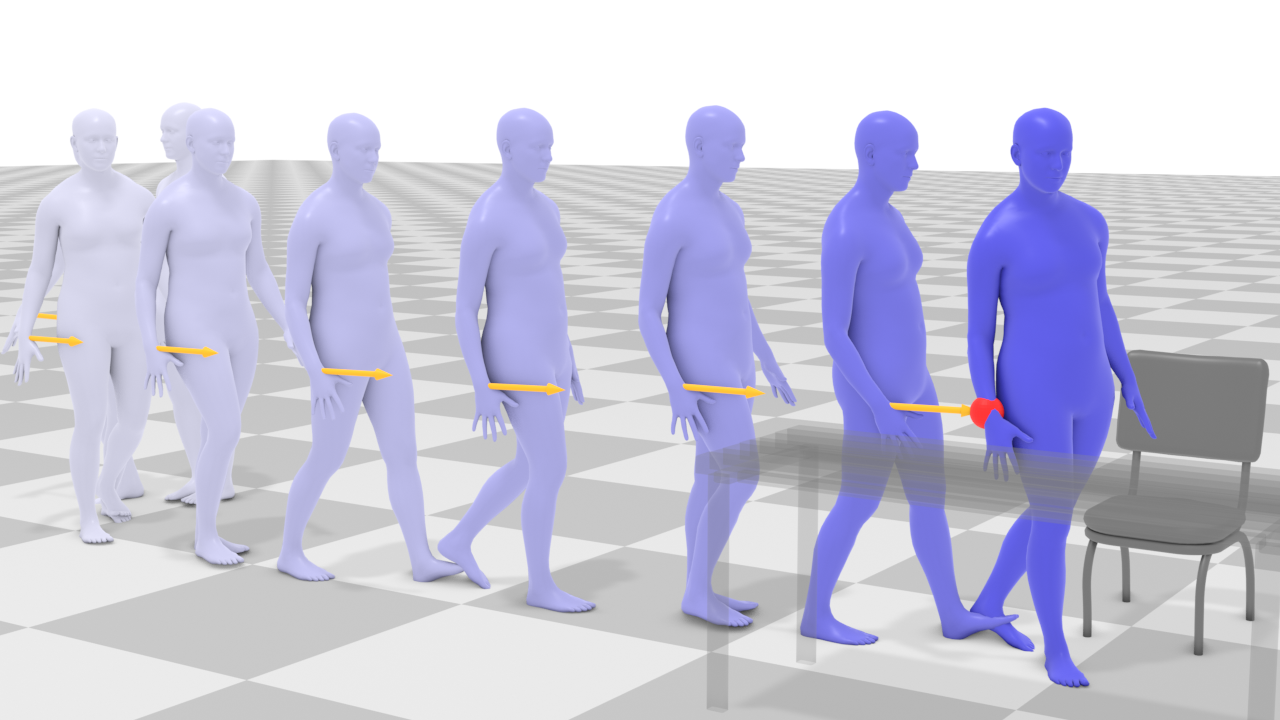}
    \caption{Single-joint Control.}
    \label{fig:single_joint}
\end{subfigure}
\begin{subfigure}[b]{0.48\linewidth}
    \centering
    \includegraphics[width=\linewidth]{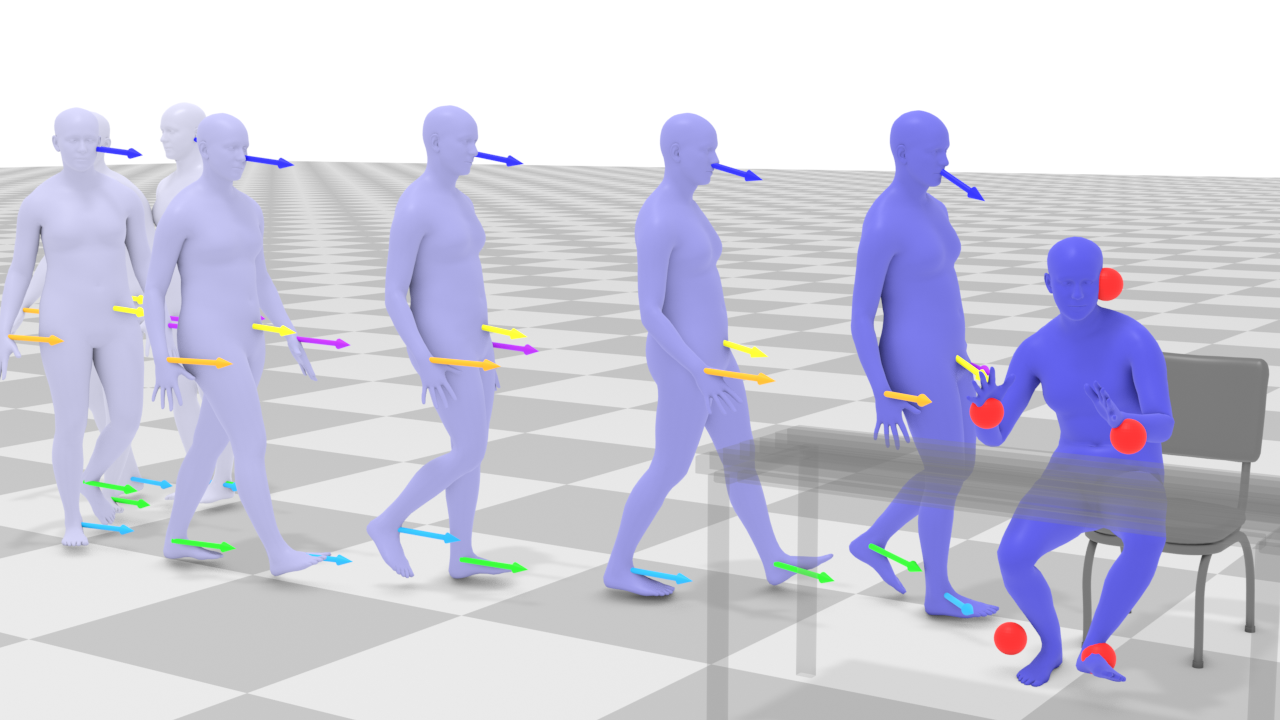}
    \caption{Multi-joint Control.}
    \label{fig:multi_joint}
\end{subfigure}

\vspace{3mm} 

\begin{subfigure}[b]{0.97\linewidth}
    \centering
    \includegraphics[width=\linewidth]{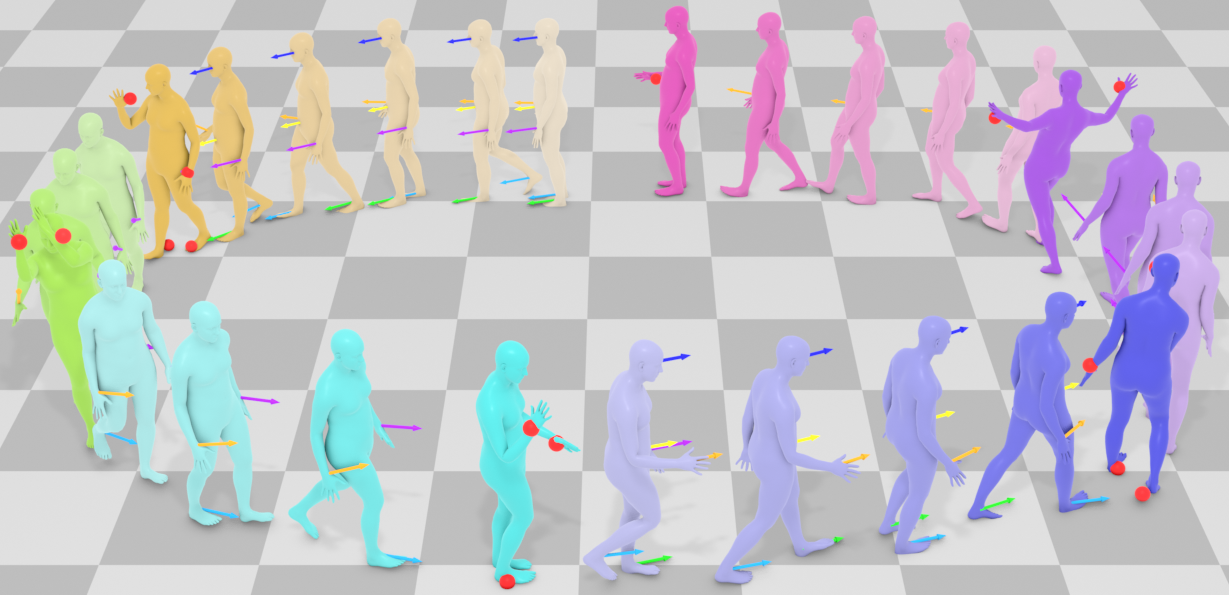}
    \caption{Long-horizon generation with arbitrary target sets.}
    \label{fig:long_horizon}
\end{subfigure}

\caption{Capabilities of the proposed COMET framework. A single model seamlessly handles both single-joint and multi-joint goal-reaching tasks while sustaining stable, realistic motion throughout long-horizon generation with arbitrary target sets.}
\label{fig:flexible_control}
\end{figure}

Human movement is inherently goal-oriented, a challenge defined not just by simple navigation but by the coordinated control of multiple joints to achieve complex target configurations, such as simultaneously positioning the pelvis, knees, and torso to achieve a sitting posture, or coordinating hands for object manipulation. This multi-joint coordination becomes even more complex when considering the temporal aspect, as the system must generate smooth and plausible transitions throughout the entire motion sequence. To address this problem, deep generative models have been studied recently. One commonly used method involves VAEs~\cite{kingma2013auto, van2017neural}. Motion generation for given constraints has been extensively explored using VAE-based approaches~\cite{ling2020character, petrovich2021actor, petrovich2022temos, zhang2023generating}. In particular, a lightweight architecture was recently introduced that not only utilizes a VAE structure but also generates poses frame-by-frame, enabling a goal-reaching method suitable for real-time applications~\cite{diomataris2024wandr}. The advent of diffusion models~\cite{ho2020denoising, song2020denoising, nichol2021improved}, which exhibit powerful and realistic generative capabilities, has prompted their widespread application in the field of human motion generation~\cite{tevet2023human}. Especially in scenarios requiring spatial constraints~\cite{karunratanakul2023guided, xie2024omnicontrol, zhao2024dartcontrol, karunratanakul2024optimizing}, explored flexible controllability for diverse, spatially constrained motion generation tasks using diffusion models. However, these existing approaches often suffer from (i) the inability to precisely control multiple joints and targets simultaneously, (ii) difficulty in maintaining stability over extended sequences without drift or degradation, and (iii) failure to meet real-time requirements.

In this work, we propose COMET (\textbf{CO}ntrollable Long-term \textbf{M}otion Generation with \textbf{E}xtended Joint \textbf{T}argets), a novel and unified framework designed to generate long-term, stable human motion with fine-grained joint-level control for real-time application. COMET is built on an auto-regressive, VAE-based architecture that uniquely integrates two key components to address critical challenges in motion synthesis. Central to our method is an adaptive attention mechanism that flexibly conditions motion generation on an arbitrary subt of joints learned during training. This allows a single trained model to be utilized for various goal-reaching tasks by freely specifying any combination of these learned joints as control targets, without the need for retraining or model modification with a loss design that encourages generalization across arbitrary joint subsets. Specifically, this flexible joint control enables direct manipulation of the target pose, allowing users to either guide specific joints to desired locations or define entire goal postures. Furthermore, to mitigate common auto-regressive issues such as error accumulation and motion drift over extended sequences, we introduce a reference-guided feedback loop. By leveraging a Gaussian Mixture Model (GMM) trained on reference poses, our method robustly guides generated motions toward a learned manifold of natural human poses, substantially enhancing realism and stability. Additionally, we extend this reference-guided feedback to a plug-and-play stylization capability, allowing users to effortlessly generate stylized locomotion by simply adjusting the reference dataset. Through the integration of a lightweight autoregressive VAE architecture with the proposed reference-guided feedback mechanism, COMET achieves stable, fine-grained, and real-time generation of controllable human motion.

COMET is designed as a versatile framework capable of addressing diverse tasks with a single model. We demonstrate its effectiveness across four primary applications: single-joint goal-reaching, goal-reaching with an arbitrary set of joints, motion in-betweening, and motion stylization. Through comprehensive experiments, we show that COMET quantitatively and qualitatively outperforms state-of-the-art models. Furthermore, we confirm its ability to generate stable motion over long horizons while successfully completing the specified tasks.

The principal contributions of our work are threefold:
\begin{itemize}
\item A flexible, Transformer-based control mechanism that enables a single model to generate motion conditioned on any arbitrary subset of target joints, supporting diverse tasks like goal-reaching and in-betweening without retraining.
\item A reference-guided feedback loop that ensures long-term temporal stability by grounding generated motion in a learned manifold of natural poses, effectively mitigating error accumulation and drift.
\item A novel plug-and-play stylization capability that extends the feedback loop, allowing motion styles to be applied and modified at runtime.
\end{itemize}

%% file: sec/2_related_work.tex
\begin{figure*}[htbp]
    \centering
    \includegraphics[width=\textwidth]{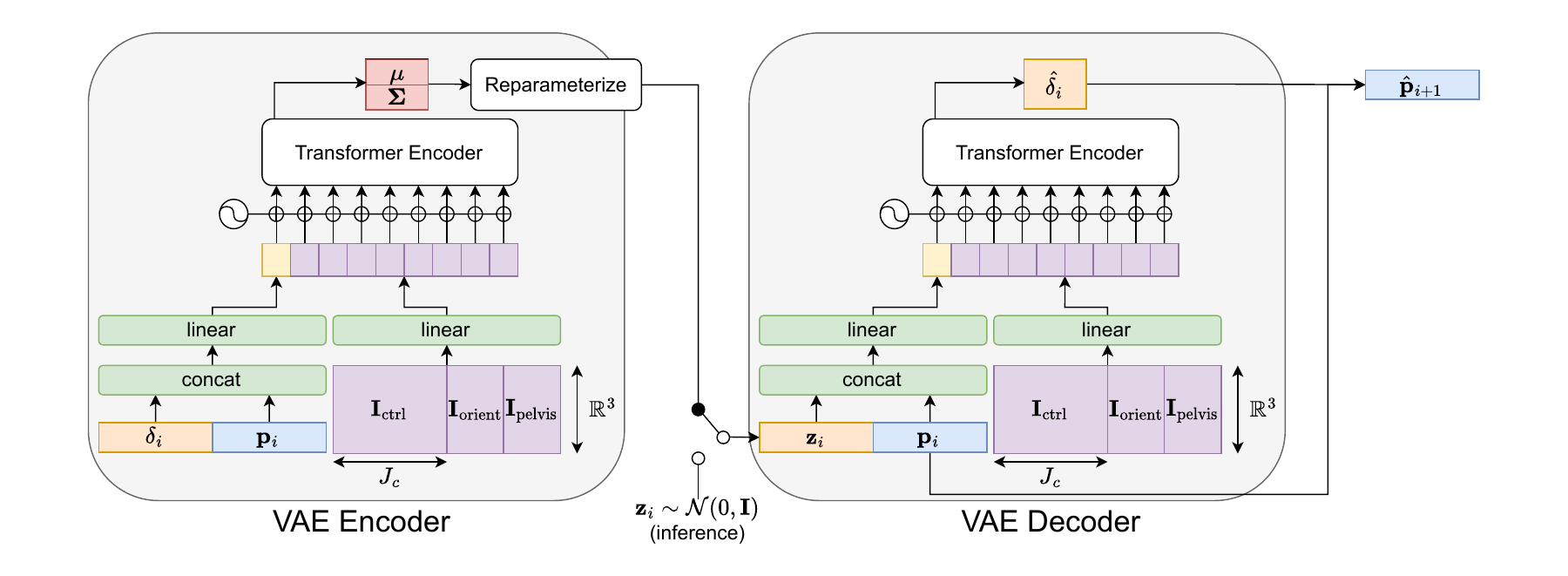}
    \caption{Architecture of COMET. The model employs a conditional Variational Autoencoder (c-VAE) with Transformer encoder layers. It processes the current state feature ($\mathbf{p}_i$), delta feature ($\mathbf{\delta}_i$), and composite intention feature ($\mathbf{I}_i$). Transformer attention is applied to intention features; It enables adaptive control by allowing the model to conditionally attend only to information from actively controlled joints based on the input controlling joint signal. The system then auto-regressively predicts the subsequent delta feature ($\hat{\mathbf{\delta}}_{i}$), which is a required change to the current pose to achieve goal-reaching task.}
    \label{fig:architecture_diagram}
\end{figure*}

\section{Related Work}
\label{sec:2_related_work}

\subsection{Controllable Motion Generation}
Controllable motion generation is crucial for enhancing the usability of motion synthesis systems by allowing users to guide the generation process. One common approach uses language-level conditioning, leveraging either action labels~\cite{guo2020action2motion, petrovich2021actor} or text descriptions~\cite{guo2022generating, zhang2023generating, athanasiou2024motionfix}. Alternatively, spatial constraints offer a more direct way to guide human motion, including guiding motions to follow a specific trajectory~\cite{karunratanakul2023guided, shafir2023priormdm} or to reach designated target joint positions~\cite{ling2020character, xie2024omnicontrol, zhao2024dartcontrol, diomataris2024wandr}. Building on the success of MDM~\cite{tevet2023human}, both PriorMDM~\cite{shafir2023priormdm} and CondMDI~\cite{cohan2024condmdi} use MDM as a prior to enable conditional motion generation. Recent advances in diffusion models have significantly expanded capabilities, including multi-joint control~\cite{xie2024omnicontrol, karunratanakul2024optimizing, zhao2024dartcontrol}. However, these diffusion-based approaches often suffer from high computational cost. For instance, methods such as~\cite{karunratanakul2024optimizing, zhao2024dartcontrol} employ optimization-based approaches for constrained motion generation, making them less suitable for real-time applications. Furthermore, because they generate motion in fixed-length segments, these methods often struggle to adapt to rapid changes in control signals and become inefficient when synthesizing long sequences.

\subsection{Long-term Motion Generation}
Synthesizing long, coherent motion sequences presents distinct challenges, and prior work has tackled this problem from two perspectives: extending sequence length with fixed-size generative models and frame-by-frame autoregressive synthesis. Non-autoregressive approaches typically generate motion in fixed-length segments and then stitch them together. For example, PriorMDM~\cite{shafir2023priormdm} uses a two-stage diffusion process to produce subsequences and then refines their boundaries for smoother transitions. Similarly, Bae et al.~\cite{bae2025less} adopt a keyframe-centric strategy, first generating sparse key poses and then interpolating in-between motions, which helps maintain global coherence across very long sequences. Another line of work leverages powerful diffusion priors with test-time optimization to achieve long-horizon control. Recent diffusion models~\cite{karunratanakul2024optimizing, zhao2024dartcontrol} can extend motion length by stitching boundary conditions; however, they require heavy computation during inference, making real-time use difficult. Reinforcement learning methods for long-term generation, such as physics-driven controllers~\cite{ling2020character, tevet2024closd}, typically demand training with external simulators and can exhibit unnatural, jittery movements. Data-driven autoregressive models present a more direct path for continuous motion synthesis. Most relevant to our work, WANDR~\cite{diomataris2024wandr} generates motion one frame at a time, enabling interactive user control over long horizons. Yet purely autoregressive synthesis tends to accumulate small errors over time, which gradually amplify and ultimately cause the synthesized motion to drift, degrade, or even collapse entirely over long sequences.

\subsection{Motion Stylization}
Motion stylization methods seek to alter the stylistic properties of a motion while preserving its essential content. Early works approached this by explicitly disentangling motion into content and style components and then recombining them. For example, Aberman et al.~\cite{aberman2020unpaired} introduced an unpaired training framework to transfer style between motions by mapping them into a shared content space, while Guo et al.~\cite{guo2024genmostyle} and Raab et al.~\cite{raab2024momo} learned separate latent representations for motion content and style that can be swapped to produce stylized motions. SMooDi~\cite{zhong2024smoodi} introduces a diffusion-based model that can generate stylized motions given a text description and a style example sequence. It employs a learned style guidance and a lightweight style adaptor to bias a pre-trained text-to-motion diffusion model toward the desired style. On the other hand, MoST~\cite{kim2024most} proposes a transformer-based architecture explicitly designed to disentangle style and content.

%% file: sec/3_method.tex
\section{Method}
\label{sec:3_Method}

Our work aims to enhance existing data-driven motion synthesis frameworks, particularly by addressing limitations in controllability and long-term stability. This section first defines the motion representation employed. Subsequently, we detail our proposed contributions: (1) Adaptive Joint Control Transformer; (2) Reference-guided Feedback (RGF) for Long-term Generation; and (3) Plug-and-Play Stylization.

\subsection{Data Representation}

We represent a motion sequence in SMPL-X~\cite{SMPL-X:2019} format as \(\mathcal{M} = \{\mathbf{p}_i\}_{i=1}^{T}\), where each per-frame state vector $\mathbf{p}_i$ is defined as the concatenation of three components: global pelvis translation, root orientation, and local joint rotations of the articulated body:
\begin{equation}
\mathbf{p}_i = \big[\, \mathbf{t}_i \; ; \; \mathbf{r}_i \; ; \; \boldsymbol{\theta}_i \,\big],
\end{equation}
where, \(\mathbf{t}_i\!\in\!\mathbb{R}^3\) is the pelvis translation in world coordinates, \(\mathbf{r}_i\!\in\!\mathbb{R}^6\) is the root orientation in 6D representation \cite{zhou2019continuity}, and \(\boldsymbol{\theta}_i\!\in\!\mathbb{R}^{J\times 6}\) are the 6D local rotations of the remaining \(J=21\) body joints. 

We also introduce a per-frame pelvis-aligned delta feature $\mathbf{\delta}_i$, computed as the change between consecutive states after removing global yaw. This delta is similarly defined as a concatenation of pelvis-plane translation, relative root orientation, and local joint rotation differences:
\begin{equation}
\mathbf{\delta}_i = \big[\, \delta \mathbf{t}_i^{xy} \; ; \; \delta \mathbf{r}_i \; ; \; \delta \boldsymbol{\theta}_i \,\big].
\end{equation}

Both state vectors $\mathbf{p}_i$ and delta vectors $\mathbf{\delta}_i$ are normalized using training-set statistics.

\begin{figure}[tbp]
\centering
\includegraphics[width=0.8\linewidth]{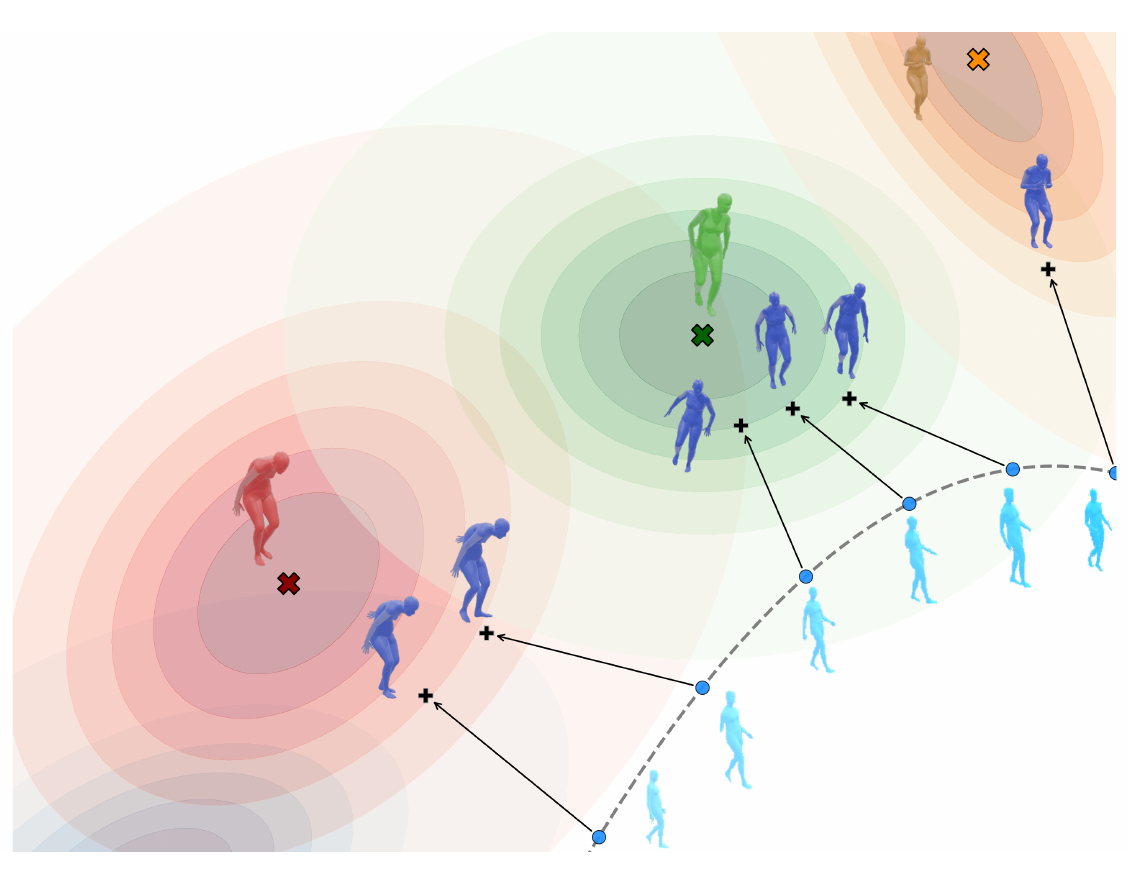}
\caption{Reference-guided feedback with GMM components as attractors. The left cluster represents the old style reference, the center cluster represents the drunken style reference, and the right cluster represents the cold style reference. The corrected pose, shown as the point moved by the arrow, is positioned relative to these references.}
\label{fig:reference-guided-style}
\end{figure}

\subsection{Adaptive Joint Control Transformer}

COMET generates motion autoregressively, predicting one frame at a time while conditioning on: (i) the current state $\mathbf{p}_i$, (ii) the delta feature $\mathbf{\delta}_i$, and (iii) a composite intention vector $\mathbf{I}_i$. Inspired by WANDR~\cite{diomataris2024wandr}, we extend the intention vector to support arbitrary sets of user-specified control joints, enabling flexible goal-directed guidance.

\paragraph{Intention Conditioning.}
Three complementary intention signals are adopted to jointly guide global joint trajectories, global body placement, and character orientation. For each controlled joint $j \in J_c$, the \emph{control joint intention} $\mathbf{I}_{\text{ctrl}}$ drives the joint toward its designated global target position $\mathbf{G}_j$:
\begin{equation}
\mathbf{I}_{\text{ctrl}, j, i} = \frac{\mathbf{G}_j - \mathbf{P}_{j,i}}{t_{\text{goal}, j} - i},
\end{equation}
where $\mathbf{P}_{j,i}$ is the joint’s current global position at frame $i$, and $t_{\text{goal}, j}$ is its target frame index. 

To stabilize overall body placement, we introduce a \emph{target-center intention} $\mathbf{I}_{\text{pelvis}}$ that attracts the pelvis toward the average XY position of all controlled joints:
\begin{equation}
\mathbf{G}_{\text{avg}, i}^{xy} = \frac{1}{|J_c|} \sum_{j \in J_c} \mathbf{G}_j^{xy},
\end{equation}
\begin{equation}
\mathbf{I}_{\text{pelvis}, i} = 2 \cdot \left(1 - e^{-\lVert \mathbf{G}_{\text{avg}, i}^{xy} - \mathbf{P}_i^{xy} \rVert_2}\right) 
\cdot \frac{\mathbf{G}_{\text{avg}, i}^{xy} - \mathbf{P}_i^{xy}}{\lVert \mathbf{G}_{\text{avg}, i}^{xy} - \mathbf{P}_i^{xy} \rVert_2},
\end{equation}
where $\mathbf{P}_i^{xy}$ denotes the current pelvis XY position. This term encourages coherent body translation even when multiple joints are being independently controlled.

Finally, the \emph{orientation intention} $\mathbf{I}_{\text{orient}}$ steers the character’s heading toward the desired global facing direction:
\begin{equation}
\mathbf{I}_{\text{orient}, i} = \mathbf{h}_{\text{goal}}^{xy} - \mathbf{h}_i^{xy},
\end{equation}
where $\mathbf{h}_i^{xy}$ is the current horizontal heading vector.

By combining these three signals, COMET achieves precise, stable, and coherent motion under arbitrary configurations of controlled joints. This design enables a single model to flexibly handle diverse tasks such as goal-reaching, trajectory following, and coordinated multi-joint manipulation.

\paragraph{Joint-wise Attention for Adaptive Joint Control}  
To generalize across arbitrary sets of controlled joints, we introduce a \textit{joint-wise key padding mask} applied to the intention tokens. For each motion sequence $\mathcal{M}$ of length $S$, we sample a binary mask vector $m \in \{0,1\}^{|J_c|}$ over the controllable joints and keep it fixed for the entire sequence. Each element $m_j = 1$ indicates that joint $j$ is actively controlled, while $m_j = 0$ disables its control and prevents its corresponding intention token from being attended to. This activation is controlled by the Transformer~\cite{vaswani2017attention}'s attention masking mechanism, ensuring that only active control joints contribute to the generation process. By varying the number of active joints across sequences, the model is encouraged to handle diverse control configurations robustly. At frame $i$, the intention vector is defined as:

\begin{equation}
\mathbf{I}_i = \big[\, \mathbf{I}_{\text{pelvis}, i} \; ; \; \mathbf{I}_{\text{orient}, i} \; ; \; \{\mathbf{I}_{\text{ctrl}, j, i}\}_{j \in J_c} \,\big],
\label{eq:intention_vector}
\end{equation}
where the first two components are always active, while the control-related tokens are selectively masked by $m$ through the Transformer's attention mechanism.

\paragraph{Generative Framework}

COMET employs a conditional Variational Autoencoder (c-VAE), with both its encoder $q_{\phi}(\mathbf{z}|\mathbf{x})$ and decoder $p_{\theta}(\mathbf{x}|\mathbf{z})$ implemented using Transformer encoder layers \cite{vaswani2017attention}. At frame $i$, the encoder input comprises a single initial token obtained by concatenating linearly-projected representations of the current state feature $\mathbf{p}_i$ and the delta feature $\mathbf{\delta}_i$. Remaining input tokens are reserved for the intention features. The Transformer encoder's first output token predicts the mean and variance of the posterior distribution of the latent variable $\mathbf{z}_i \sim q_{\phi}(\mathbf{z}_i|\mathbf{p}_i,\mathbf{\delta}_i,\mathbf{I}_i)$. The VAE decoder also uses Transformer encoder architecture; however, instead of delta features, the latent vector $\mathbf{z}_i$ is utilized as input along with state and intention features. The decoder then predicts delta features at its first output token. Consequently, after training, the decoder can generate next-frame pose deltas conditional on current body state, intention features, and latent noise. We set $J_c$ to be a total of six controlling joints by adding the pelvis to five end-effectors. The overall architecture of COMET and its data flow are illustrated in Figure~\ref{fig:architecture_diagram}.

\paragraph{Loss Function}
The COMET framework is trained end-to-end by minimizing a total loss function $\mathcal{L}_{total}$, which comprises three main components: a reconstruction loss for the delta feature, a Kullback-Leibler (KL) divergence loss for the VAE regularization, and a joint position loss to ensure accurate control of specified joints.

The reconstruction loss measures the Mean Squared Error (MSE) between the ground truth delta feature $\mathbf{\delta}_i$ and the predicted delta feature $\hat{\mathbf{\delta}}_i$ from the decoder to learn pose deltas:

\begin{equation}
\mathcal{L}_{recon} = \mathbb{E} [ \lVert \mathbf{\delta}_i - \hat{\mathbf{\delta}}_i \rVert_2^2 ].
\end{equation}

The KL divergence loss is a standard VAE component that regularizes the latent space by encouraging the learned distribution $q_{\phi}(\mathbf{z}_i | \mathbf{p}_i, \mathbf{\delta}_i, \mathbf{I}_i)$ to be close to a standard Gaussian prior $\mathcal{N}(\mathbf{0},\mathbf{I})$:

\begin{equation}
\mathcal{L}_{KL} = D_{KL}(q_{\phi}(\mathbf{z}_i | \mathbf{p}_i, \mathbf{\delta}_i, \mathbf{I}_i) || \mathcal{N}(\mathbf{0}, \mathbf{I})).
\end{equation}

The joint control loss supervises the accuracy of the controlled joints. It is defined as the MSE between the ground truth global positions of the joints in the control set $J_c$ at the next frame ($i+1$) and their predicted positions. 

\begin{equation}
\mathcal{L}_{joint} = \mathbb{E}\left[\frac{1}{|J_c|}\sum_{j \in J_c} \lVert \mathbf{P}_{j,i+1}-\hat{\mathbf{P}}_{j,i+1}\rVert_2^2\right].
\end{equation}

The total loss is a weighted sum of these components:

\begin{equation}
\mathcal{L}_{total} = \mathcal{L}_{recon} + \lambda_{KL} \mathcal{L}_{KL} + \lambda_{joint} \mathcal{L}_{joint},
\end{equation}
where $\lambda_{KL}$ and $\lambda_{joint}$ are hyperparameter weights that balance the contribution of each term.

\subsection{Reference-guided Feedback for Long-term Generation}
\label{sec:reference_based_feedback}

To ensure long-term stability, we guide generated poses toward a manifold of natural human motion by leveraging a compact set of reference poses. The reference pose distribution is modeled with a Gaussian Mixture Model (GMM), enabling smooth and data-driven corrections during inference.

We first construct a reference motion dataset, which may contain multiple motion sequences or even a single exemplar sequence. Feature vectors $\mathbf{f} = [\mathbf{r}, (\mathbf{t})_z]$, comprising rotational component and pelvis height, are extracted and used to train a GMM, where the number of mixture components $K$ is a hyperparameter. Each $k$-th Gaussian component represents a characteristic pose configuration from the reference set.

During inference, the VAE decoder predicts a pose delta, which is added to the previous pose to produce the current estimate $\hat{\mathbf{p}}_{i+1}$. To identify the most compatible reference configuration, we compute the Mahalanobis distance between the feature $\hat{\mathbf{f}}_{i+1} = [\hat{\mathbf{r}}_{i+1}, (\hat{\mathbf{t}}_{i+1})_z]$ and the mean of each Gaussian mixture $\boldsymbol{\mu}_k$. The closest component $k^*$ is selected, and the generated pose is corrected as:

\begin{equation}
\mathbf{f}_{i+1} = \hat{\mathbf{f}}_{i+1} + \alpha(\boldsymbol{\mu}_{k^*} - \hat{\mathbf{f}}_{i+1}),
\end{equation}
where $\alpha$ controls the strength of the correction. To prevent interference with precise target reaching, reference-guided feedback is disabled when the pose is near the final goal.

This mechanism offers a lightweight yet powerful solution to counteract error accumulation, analogous to ground-truth feedback but without requiring an external supervisory framework such as reinforcement learning. By continuously nudging predictions toward the learned motion manifold, our method maintains realism and stability throughout long-horizon generation.

\subsection{Plug-and-Play Stylization}

The reference-guided feedback mechanism naturally extends to plug-and-play stylization without modifying the core COMET architecture. By training the GMM on motions exhibiting a particular style, the generated poses are guided toward the stylistic motion manifold during inference. For example, a GMM trained on human motions mimicking dinosaur-like walking encourages the model to synthesize motions with distinctive dinosaur-inspired postures. Because the style information is entirely encapsulated in the GMM, real-time style switching can be achieved by simply swapping the active GMM at inference time. This decoupled design allows COMET to generate diverse stylized motions without retraining or fine-tuning, offering a flexible framework for applications. Figure~\ref{fig:reference-guided-style} illustrates this process.

%% file: sec/4_experiments.tex
\section{Experiments}
\label{sec:4_experiments}

\begin{table}[tbp]
\centering
\small
\setlength{\tabcolsep}{2pt} 
\caption{Ablations on single-joint control (right wrist). JA = Joint-wise Attention, RGF = Reference-Guided Feedback.\protect\footnotemark}
\label{tab:single-joint-task}
\begin{tabular*}{\columnwidth}{@{\extracolsep{\fill}}lccccc@{}}
\toprule
Method & JA & RGF & SR ($\uparrow$) & FS ($\downarrow$) & DTG (cm) ($\downarrow$) \\
\midrule
\multicolumn{6}{@{}l}{\textit{Baselines}} \\
GOAL~\cite{taheri2022goal}       &   &   & 0  & 29 & 149.2 \\
WANDR~\cite{diomataris2024wandr} &   &   & 32 & 16 & 24.8 \\
WANDR~\cite{diomataris2024wandr} +RGF  &   & \cmark & 49.5 & \textbf{8.9} & 21.7 \\
\addlinespace[2pt]
\midrule
\multicolumn{6}{@{}l}{\textit{Ours (Ablations)}} \\
COMET w/o JA                      &   & \cmark & 8.0  & 9.0  & 107.2 \\
COMET w/o RGF                      & \cmark &   & 31.5 & 19.0 & 59.3 \\
\textbf{COMET (full)}             & \cmark & \cmark & \textbf{52.7} & 14.0 & \textbf{20.3} \\
\bottomrule
\end{tabular*}
\end{table}

\begin{table}[tbp]
\centering
\small
\setlength{\tabcolsep}{0pt} 
\caption{Evaluation results on multi-joint control. Metrics are averaged over all possible joint combinations for each control set.}
\label{tab:multi-joint-avg}
\begin{tabular*}{\columnwidth}{@{\extracolsep{\fill}}cccc@{}}
\toprule
\# Ctrl. Joints & SR ($\uparrow$) & FS ($\downarrow$) & DTG (cm) ($\downarrow$) \\
\midrule
1 joint       & 71.62 \% & 13.70 \% & 16.32 \\
2 joints      & 70.41 \% & 14.94 \% & 11.67 \\ 
3 joints      & 71.16 \% & 15.04 \% & 9.81 \\ 
4 joints      & 69.42 \% & 15.36 \% & 9.79 \\
5 joints      & 68.75 \% & 15.48 \% & 9.40 \\ 
6 joints      & 68.87 \% & 15.32 \% & 9.08 \\
\bottomrule
\end{tabular*}
\end{table}

\begin{figure}[tbp]
\centering
  \begin{subfigure}[b]{0.48\linewidth}
    \includegraphics[width=\linewidth]{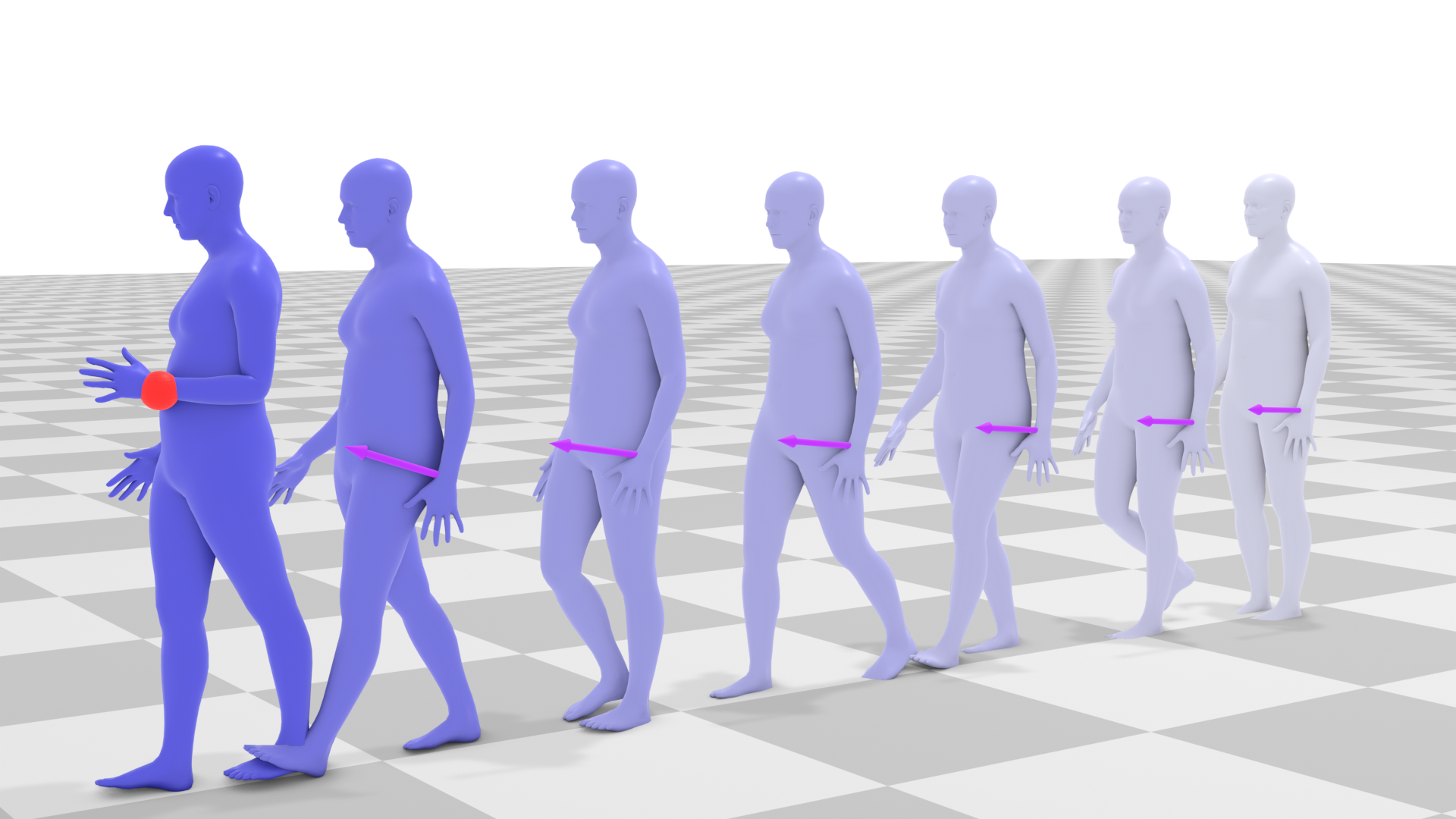}
    \caption{Single Joint Control}
    \label{fig:fig2_a_sub}
  \end{subfigure}
  \begin{subfigure}[b]{0.48\linewidth}
    \includegraphics[width=\linewidth]{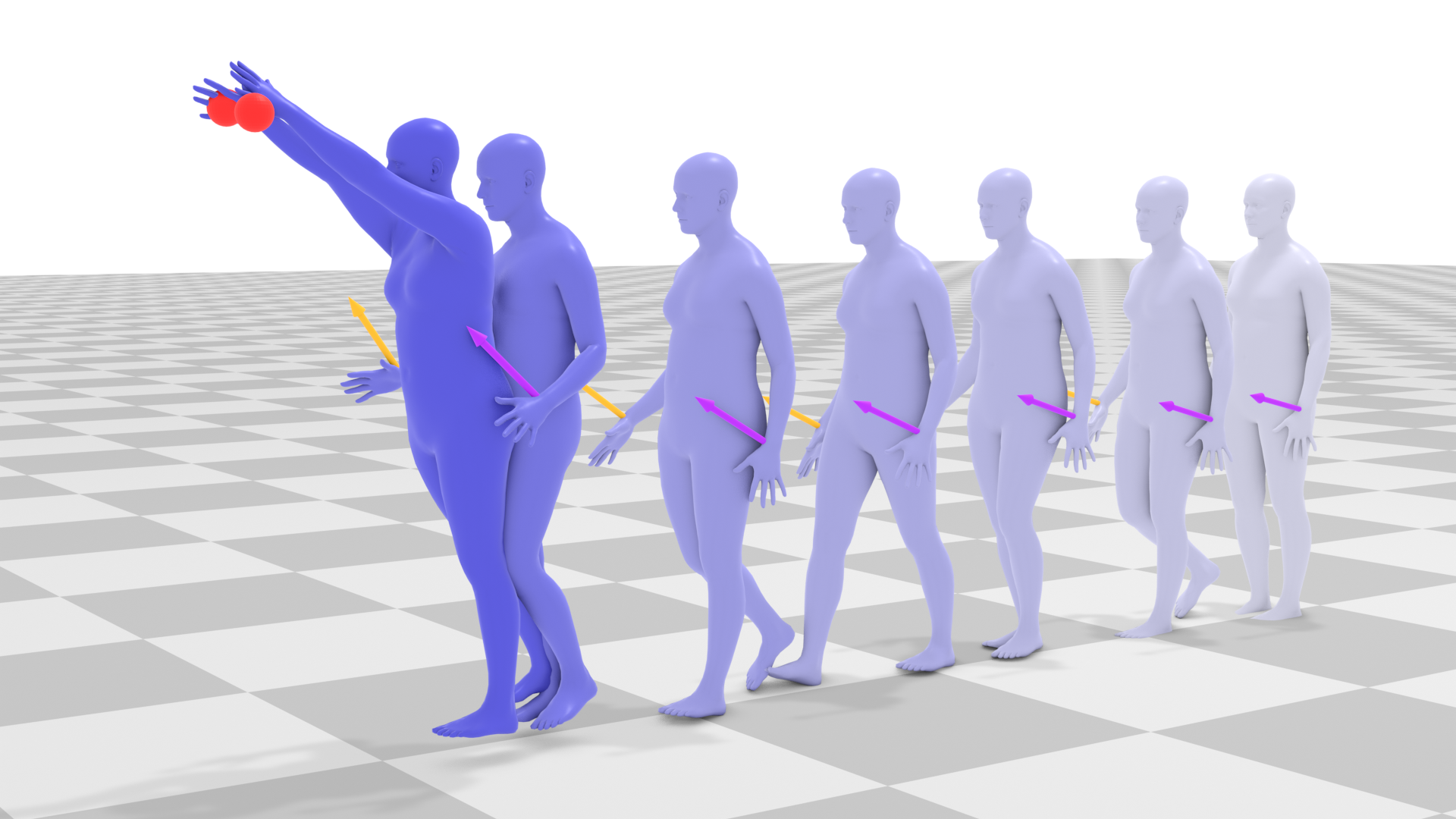}
    \caption{Two Joints Control}
    \label{fig:fig2_b_sub}
  \end{subfigure}

  \begin{subfigure}[b]{0.48\linewidth}
    \includegraphics[width=\linewidth]{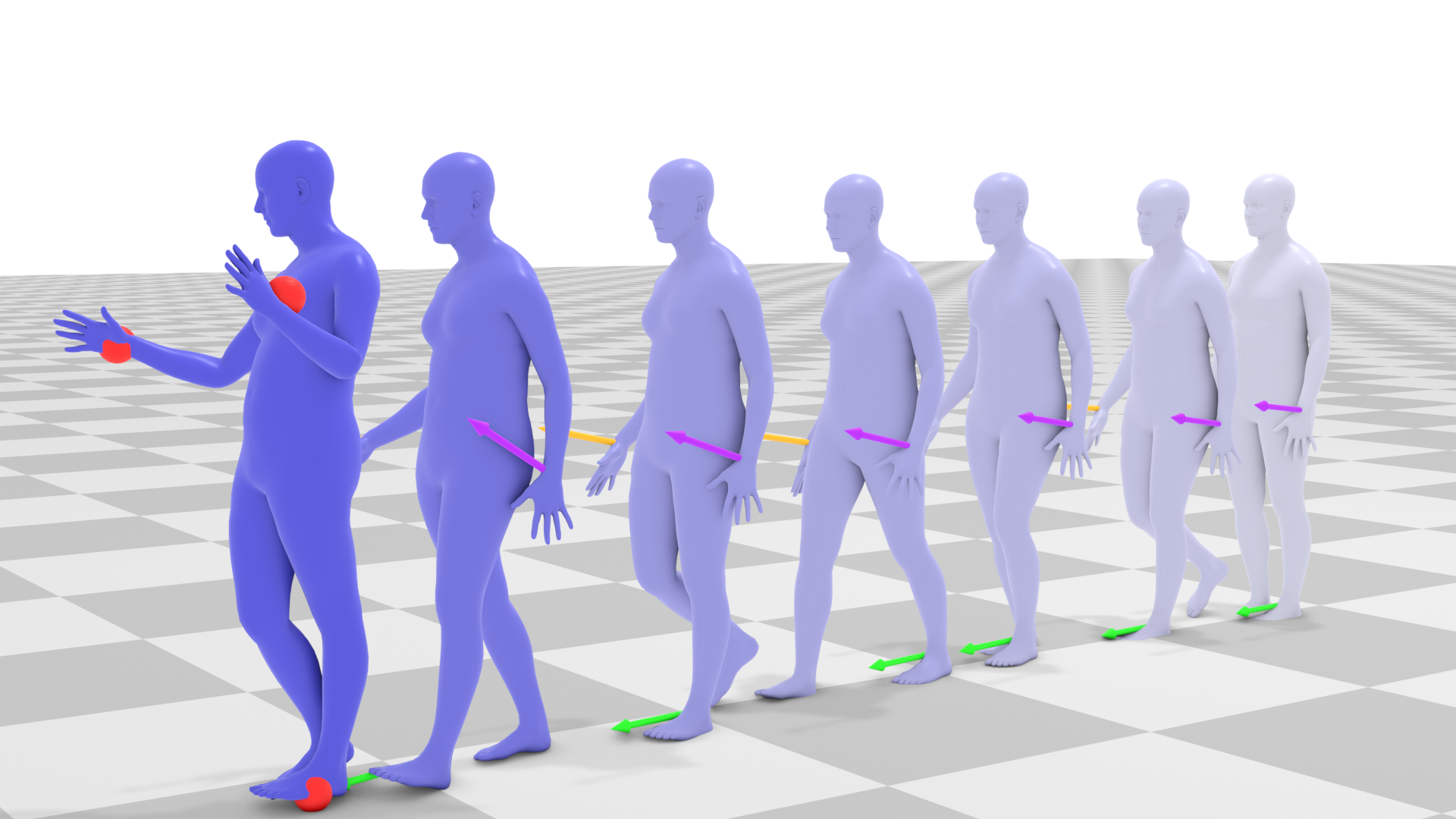}
    \caption{Three Joints Control} 
    \label{fig:fig2_c_sub}
  \end{subfigure}
  \begin{subfigure}[b]{0.48\linewidth}
    \includegraphics[width=\linewidth]{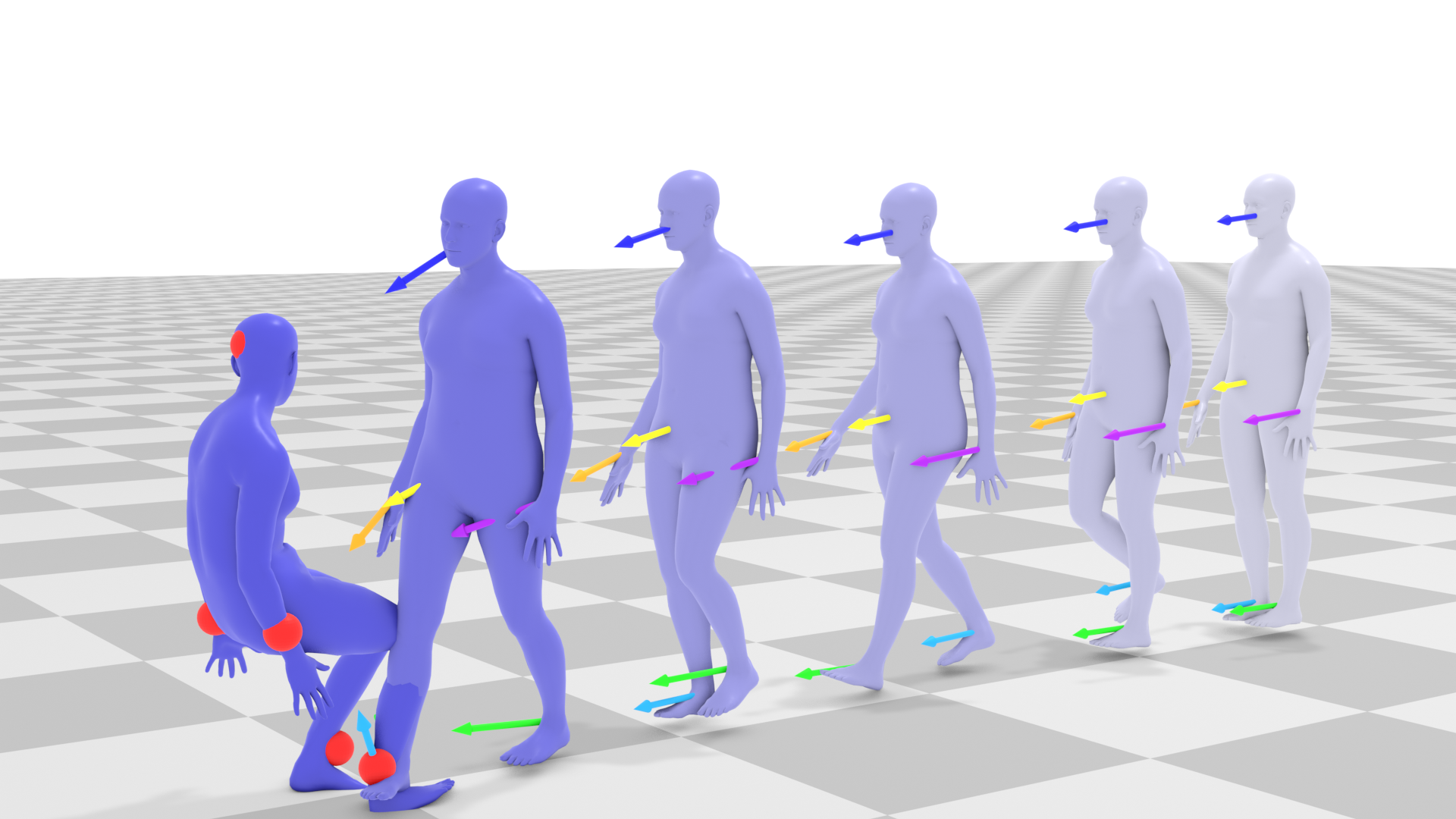} 
    \caption{Six Joints Control} 
    \label{fig:fig2_d_sub} 
  \end{subfigure}

  \caption{Qualitative Results on Multi Joint Control}
  \label{fig:multi-joint-control}
\end{figure}

We present a comprehensive evaluation establishing COMET as a versatile model for task-driven human motion. Sec.~\ref{subsec:single-multi} evaluates precise spatial control for single- and multi-joint targets; Sec.~\ref{subsec:sequential} stresses long-horizon sequential goal-reaching; and Sec.~\ref{subsec:inbetween-style} demonstrates versatility on motion in-betweening and plug-and-play stylization, including a user study. Our model is trained on AMASS~\cite{AMASS:ICCV:2019}, which contains diverse motion sequences, and CIRCLE~\cite{araujo2023circle}, which provides goal-directed sequences annotated with hand targets.

\footnotetext{COMET is designed to control up to multiple joints, but for this task, only the right wrist is activated to ensure single-joint control. The same model is used as in the multi-joint control experiments.}

\begin{table}[tbp]
\centering
\small
\setlength{\tabcolsep}{0pt}
\caption{Evaluation of long-term sequential motion generation (right wrist control).}
\label{tab:long-term-generation}
\begin{tabular*}{\columnwidth}{@{\extracolsep{\fill}}lcccc@{}}
\toprule
Goal & Method & SR ($\uparrow$) & FS ($\downarrow$) & DTG (cm) ($\downarrow$)\\
\midrule
\multirow{2}{*}{Goal 1}
& WANDR~\cite{diomataris2024wandr} & 56.16 \% & 16.57 \% & 23.56\\
& COMET & \textbf{91.55} \% & \textbf{15.29} \% & \textbf{7.22} \\
\midrule
\multirow{2}{*}{Goal 2}
& WANDR~\cite{diomataris2024wandr} & 37.33 \% & 28.18 \% & 51.51 \\
& COMET & \textbf{86.69} \% & \textbf{18.25} \% & \textbf{14.54}  \\
\midrule
\multirow{2}{*}{Goal 3}
& WANDR~\cite{diomataris2024wandr} & 20.96 \% & 39.96 \% & 99.07 \\
& COMET & \textbf{86.27} \% & \textbf{20.18} \% & \textbf{13.62}   \\
\bottomrule
\end{tabular*}
\end{table}

\begin{figure}[tbp]
  \centering 

  \begin{subfigure}[b]{\linewidth} 
    \centering 
    \includegraphics[width=\linewidth]{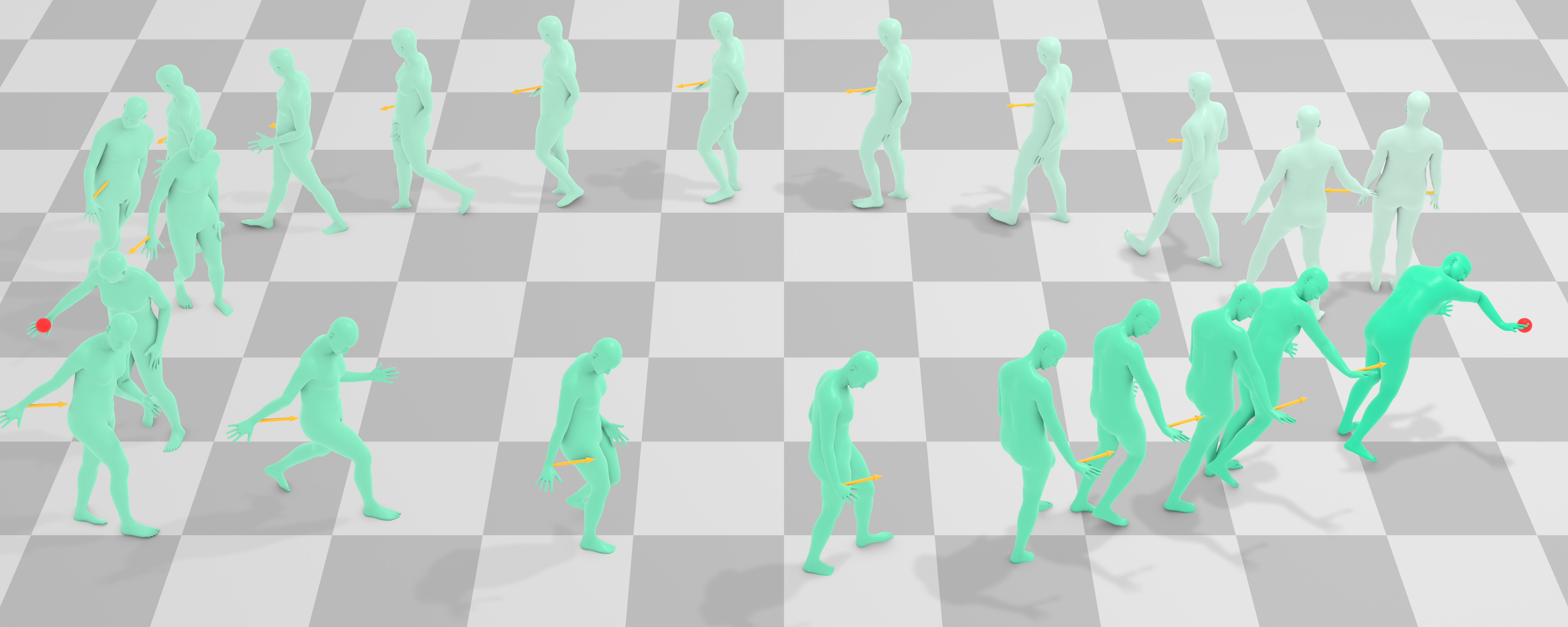}
    \caption{Without Reference-guided Feedback}
    \label{fig:wo-ref}
  \end{subfigure}
  
  \begin{subfigure}[b]{\linewidth} 
    \centering 
    \includegraphics[width=\linewidth]{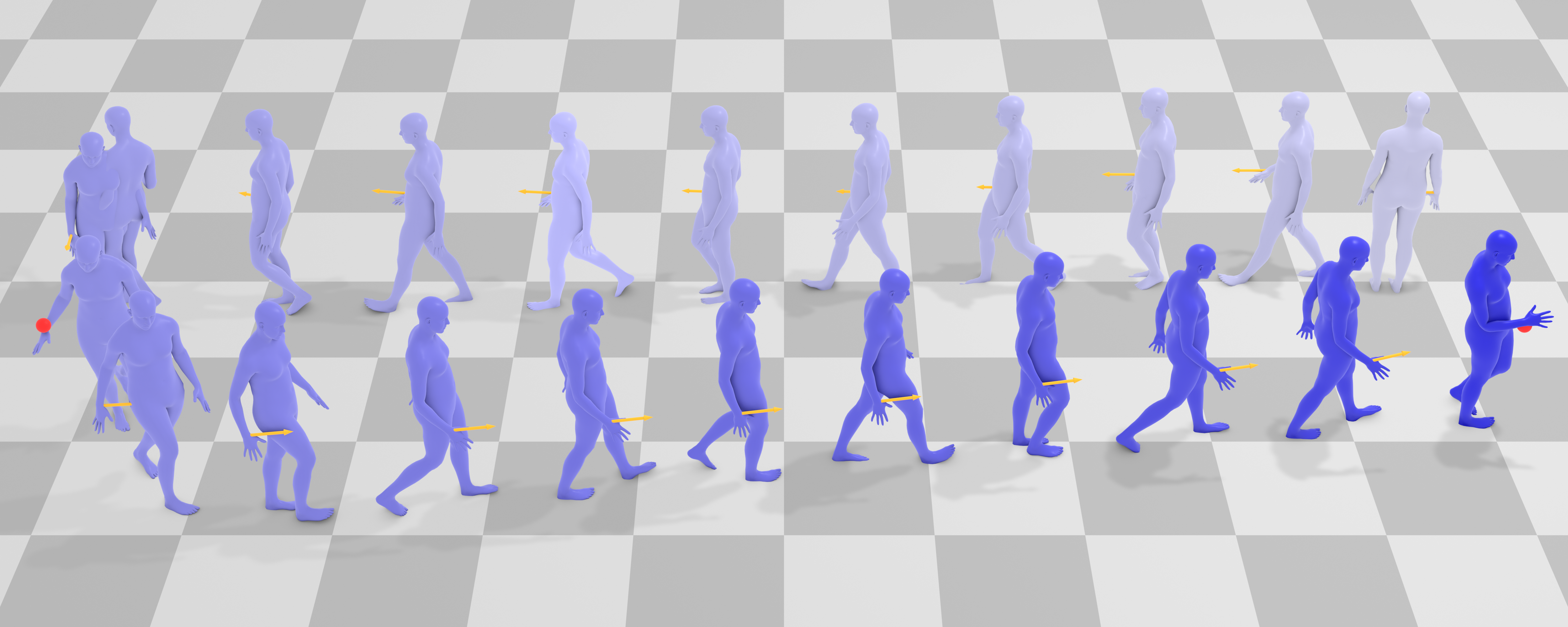}
    \caption{With Reference-guided Feedback}
    \label{fig:with-ref}
  \end{subfigure}

  \caption{Qualitative comparison of long-term sequential goal-reaching with and without the proposed Reference-Guided Feedback (RGF). (a) Without RGF, the motion rapidly diverges, exhibiting unnatural drifts and eventual collapse over extended horizons. (b) With RGF enabled, COMET maintains coherent trajectories while accurately reaching all sequential targets.}
  \label{fig:long-term-generation}
\end{figure}

\subsection{Single and Multi-Joint Control}
\label{subsec:single-multi}

\paragraph{Metrics}
We adopt a suite of quantitative measures established by \cite{diomataris2024wandr}. Our evaluation focuses on three key metrics. First, Success Rate (SR) determines the proportion of instances in which the designated target joints successfully reach within a 10-centimeter radius of their assigned 3D objectives. Second, we assess Foot Skating (FS), which quantifies the naturalness of the generated movements by measuring the degree of unrealistic foot sliding. Finally, the Distance to Goals (DTG) metric evaluates guidance accuracy by recording the smallest distances observed between the character’s target joints and their targets throughout the animation.

\begin{table}[tbp]
\centering
\small
\setlength{\tabcolsep}{0pt}
\caption{Evaluation Results on Motion In-betweening Metrics. The term `steps' refers to the number of optimization iterations performed by DNO-MDM during test-time refinement.}
\label{tab:motion-metrics}
\begin{tabular*}{\columnwidth}{@{\extracolsep{\fill}}lcccc@{}}
\toprule
Method & L2P ($\downarrow$) & L2Q ($\downarrow$) & NPSS ($\downarrow$) & FS ($\downarrow$) \\
\midrule
CondMDI~\cite{cohan2024condmdi} & 3.396 & 2.148 & 8.991 & 0.255 \\
DNO-MDM~\cite{karunratanakul2024optimizing} (100 steps) & 4.278 & 2.526 & 11.072 & 0.337 \\
DNO-MDM~\cite{karunratanakul2024optimizing} (300 steps) & 3.503 & 2.255 & 10.420 & 0.416 \\
COMET & \textbf{3.323} & \textbf{2.126} & \textbf{8.650} & \textbf{0.166} \\
\bottomrule
\end{tabular*}
\end{table}

\paragraph{Single-Joint Control}

This experiment evaluates the model’s ability to perform precise goal-reaching, where the task is to guide the right wrist to a specified 3D target location. Following the protocol of WANDR~\cite{diomataris2024wandr}, targets are uniformly sampled within a cylindrical volume with varying distance, direction, and height. Each trial begins from an unseen initial pose, requiring the model to synthesize a plausible motion sequence that reaches the goal without manual intervention. For this setting, we use the COMET model configured with six controllable joints but activate only the right wrist, with all other joints disabled, to ensure single-joint control. As shown in Table~\ref{tab:single-joint-task}, COMET achieves substantial gains across all metrics, markedly increasing the success rate (SR) while simultaneously lowering foot skating (FS) and distance-to-goal (DTG). The ablations further reveal the complementary roles of our proposed modules. COMET w/o JA uses only conditioning vectors to indicate the controlled joints but lacks the structured joint-wise attention mechanism, leading to unstable and incoherent control signals. Similarly, disabling the Reference-Guided Feedback (RGF) leads to unstable trajectories and higher drift, confirming its importance for long-horizon stability. To further validate the generality of RGF, we apply it to the baseline WANDR model without modifying its architecture. As shown in Table~\ref{tab:single-joint-task}, WANDR~\cite{diomataris2024wandr}+RGF significantly improves SR and FS over the original WANDR~\cite{diomataris2024wandr}, demonstrating that RGF is a lightweight, plug-and-play module that enhances long-term stability even for existing motion generation frameworks.

\begin{figure}[tbp]
  \centering 
  \begin{subfigure}[b]{0.8\linewidth} 
    \centering 
    \includegraphics[width=\linewidth]{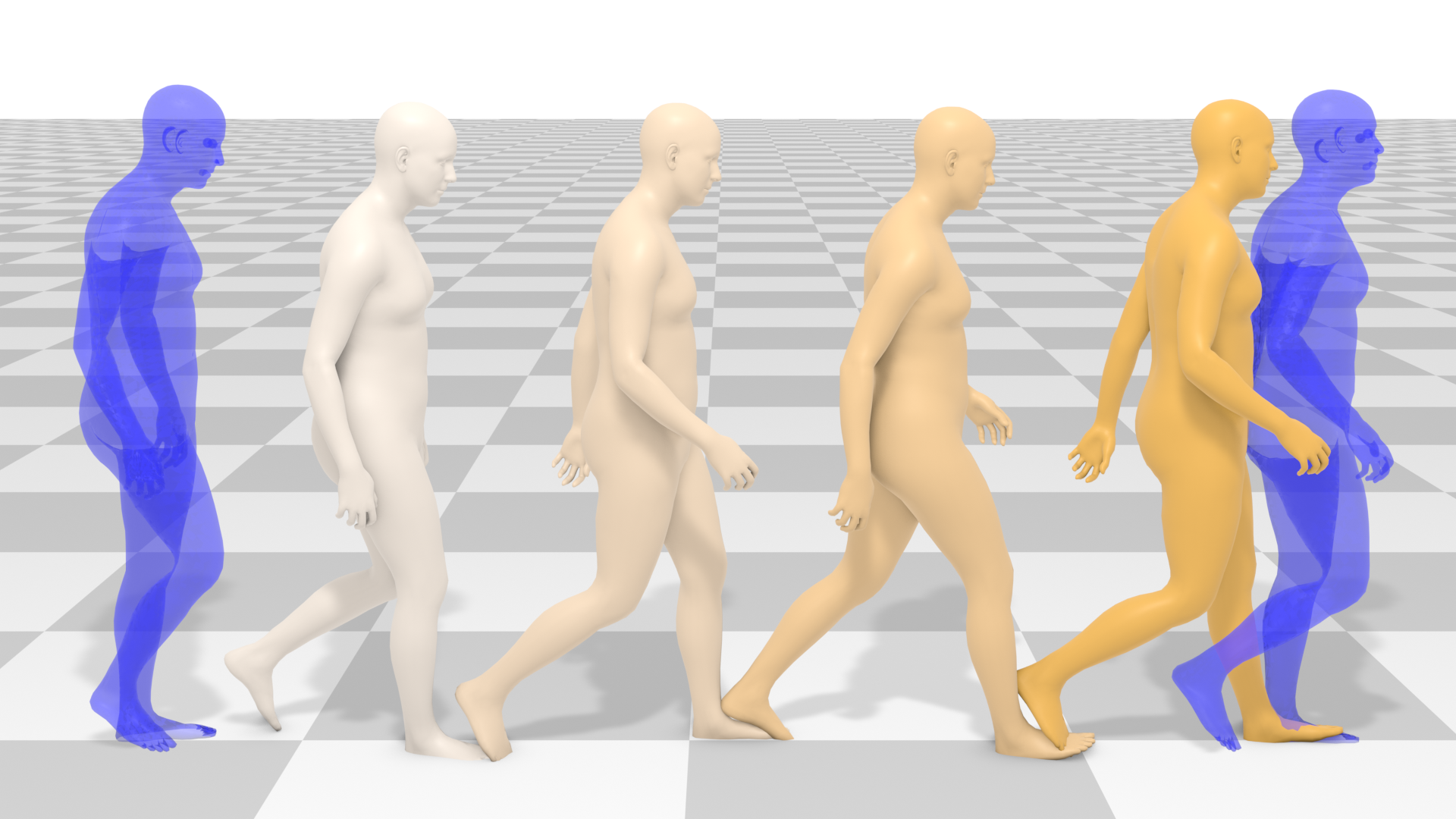}
    \caption{motion in-betweening - fast walk}
    \label{fig:mib_fig1}
  \end{subfigure}
  \begin{subfigure}[b]{0.8\linewidth} 
    \centering 
    \includegraphics[width=\linewidth]{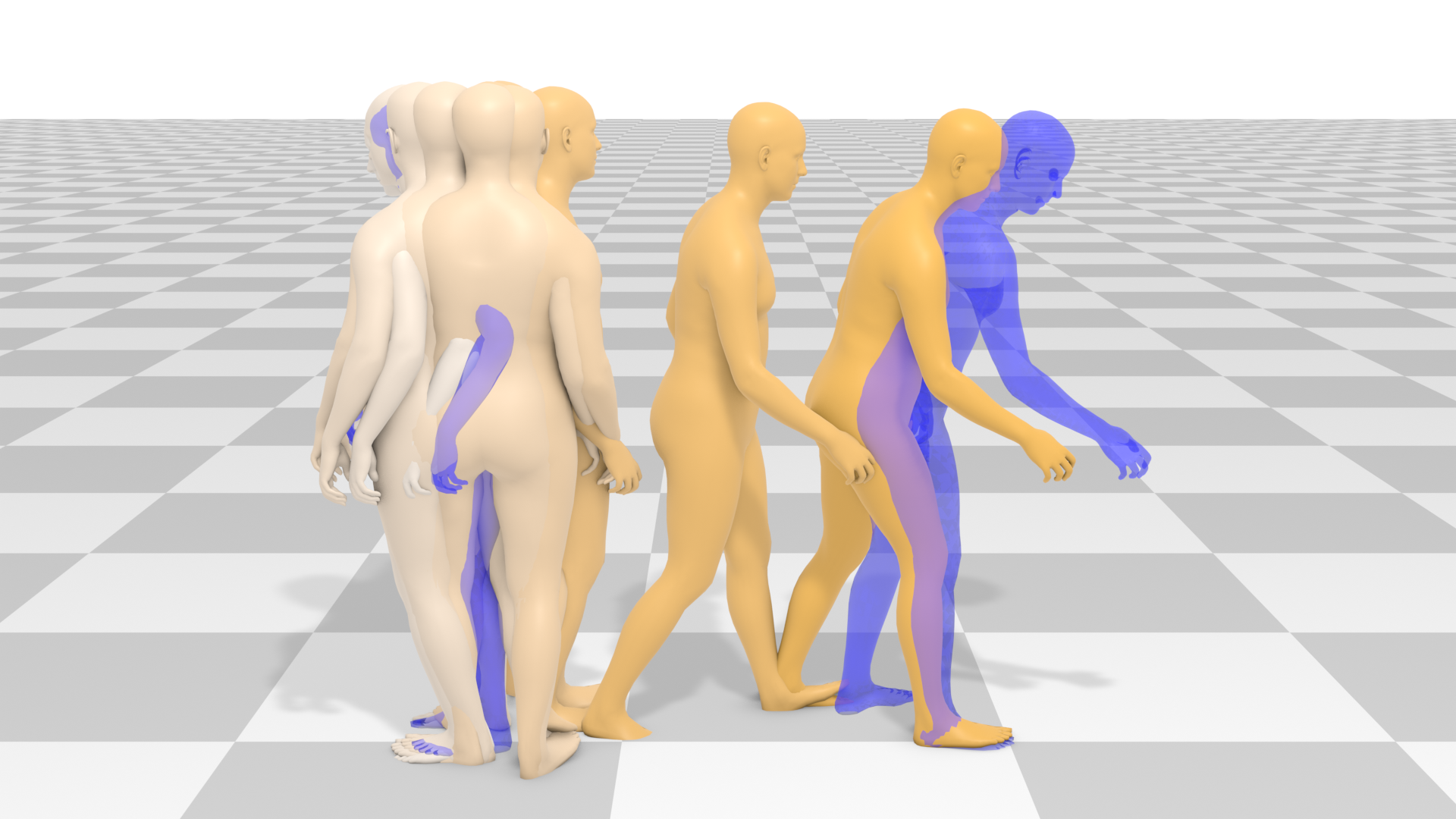}
    \caption{motion in-betweening - extend right hand}
    \label{fig:mib_fig2}
  \end{subfigure}
  \hfill 
  \begin{subfigure}[b]{0.8\linewidth} 
    \centering 
    \includegraphics[width=\linewidth]{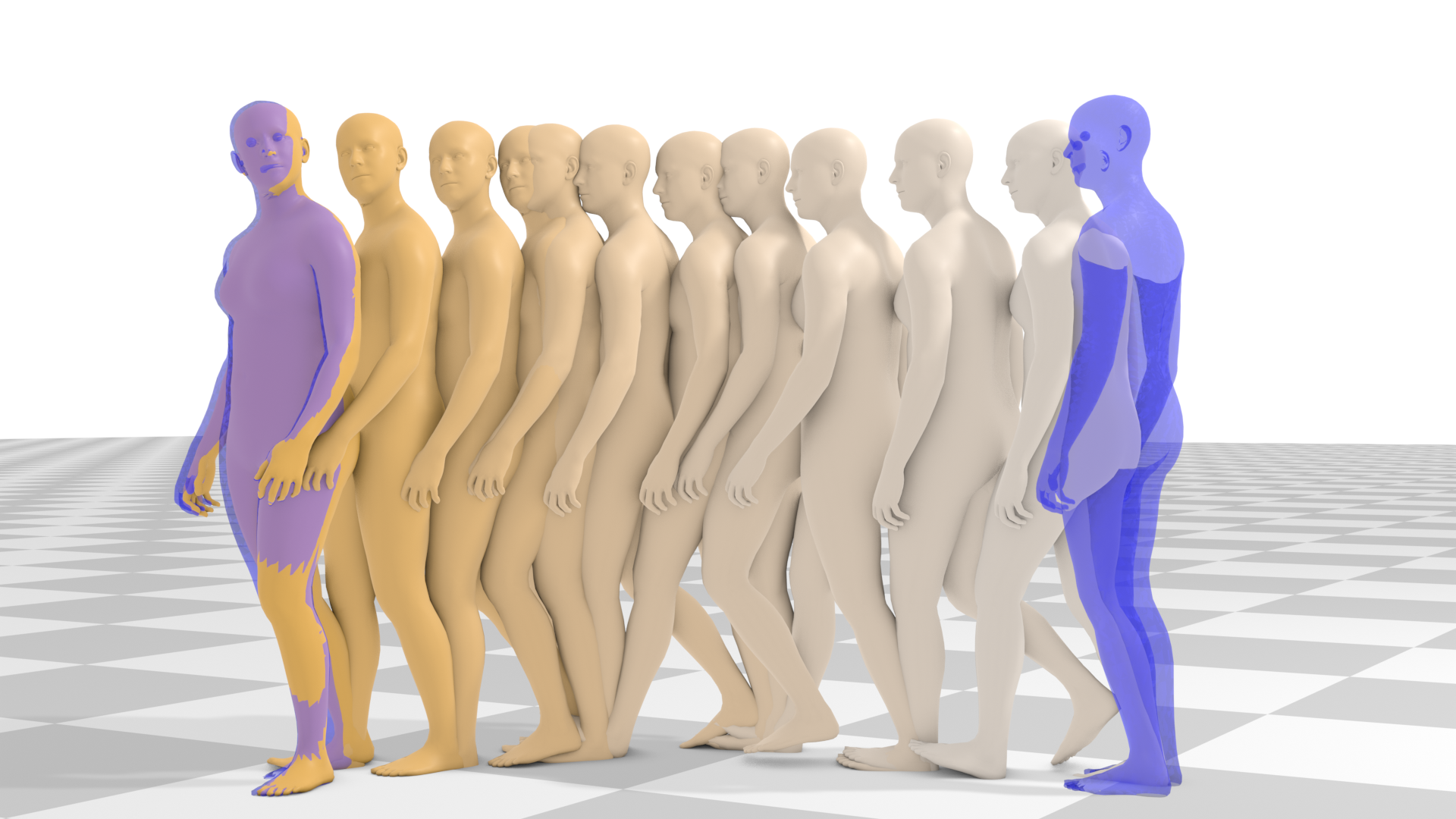}
    \caption{motion in-betweening - slow walk}
    \label{fig:mib_fig3}
  \end{subfigure}
  \caption{Motion In-betweening via Reference-guided Feedback}
  \label{fig:mib}
\end{figure}

\paragraph{Multi-Joint Control}
We evaluate arbitrary subsets of controllable joints. To resolve whole-body orientation ambiguity, we provide the desired final body orientation as input in addition to the Single-Joint Control setting. For each cardinality, we average the metrics over all possible joint combinations. Results in Table~\ref{tab:multi-joint-avg} show that DTG steadily decreases even as more joints are controlled. This reflects the effectiveness of our Joint-wise Attention, which attends to diverse joint features to extract cross-joint cues, enabling COMET to handle the increased control complexity with stable and precise motion generation. Figure~\ref{fig:multi-joint-control} shows qualitative results. Unlike the single-joint setting, combinations of multiple joints must correspond to physically realizable configurations. Therefore, we randomly sample poses and the number of controlled joints from the test dataset, then construct valid joint combinations to define the target joint positions for the quantitative evaluation.

\subsection{Long-Term Sequential Joint Control}
\label{subsec:sequential}

We evaluate sustained goal-directed locomotion by requiring models to sequentially reach a series of three targets. Each target is uniformly sampled within a cylindrical volume, and upon reaching a target, the cylinder is re-centered to determine the next goal. This task is designed to evaluate the model’s ability to maintain stability over long distances, with each target defined as a right wrist position fixed at a height of 1\,m. 
As shown in Table~\ref{tab:long-term-generation}, COMET consistently achieves substantial improvements in SR, FS, and DTG, maintaining high-quality performance regardless of sequence length or horizon complexity. 
Figure~\ref{fig:long-term-generation} illustrates the qualitative impact of the reference-guided feedback mechanism. In its absence, the generated motion sequence collapses abruptly and fails to recover.

\subsection{Motion In-betweening and Motion Stylization}
\label{subsec:inbetween-style}

\begin{figure}[tbp]
\centering

\begin{subfigure}[b]{0.48\linewidth}
    \centering
    \includegraphics[width=\linewidth]{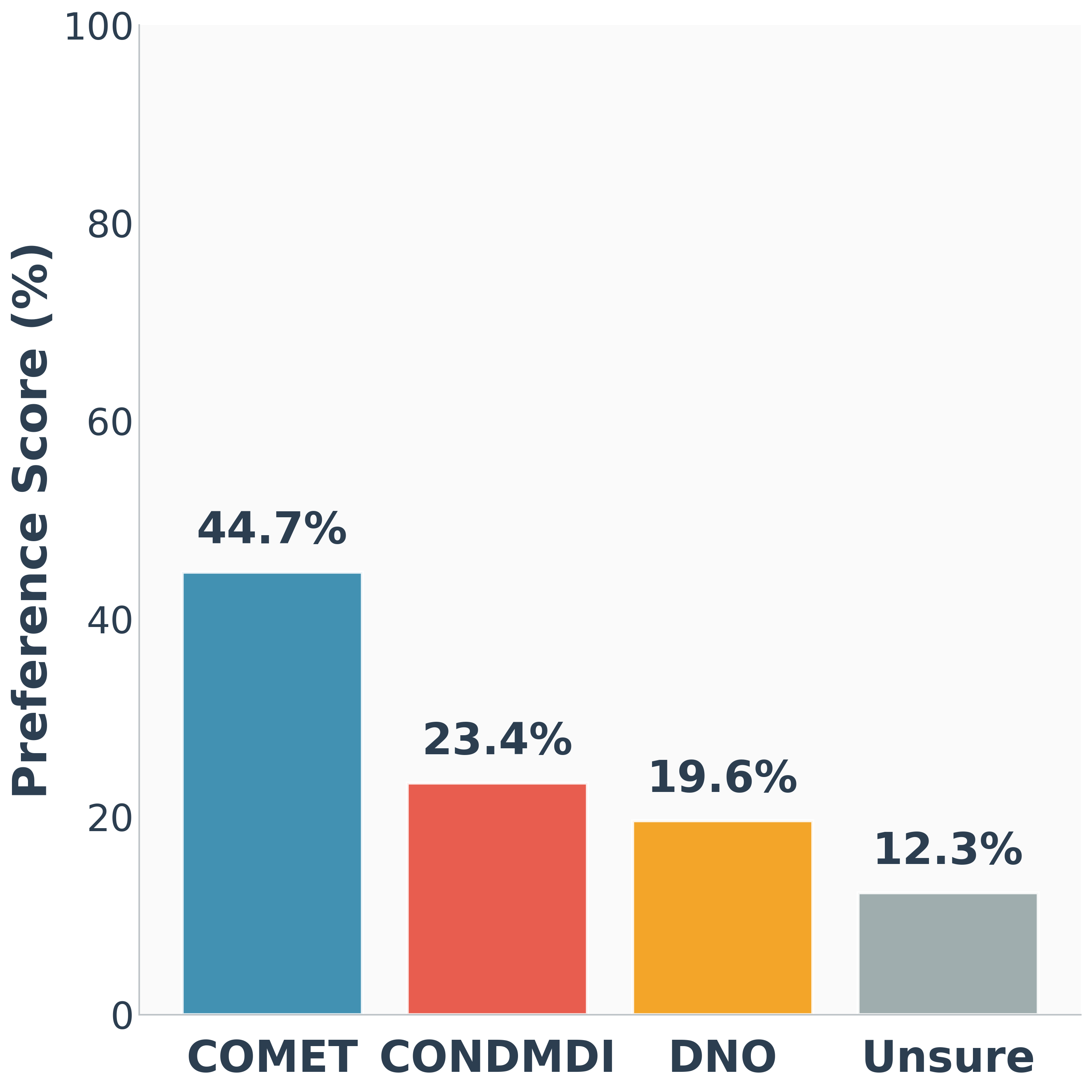}
    \caption{Motion In-betweening}
    \label{fig:mib_user_study}
\end{subfigure}
\begin{subfigure}[b]{0.48\linewidth}
    \centering
    \includegraphics[width=\linewidth]{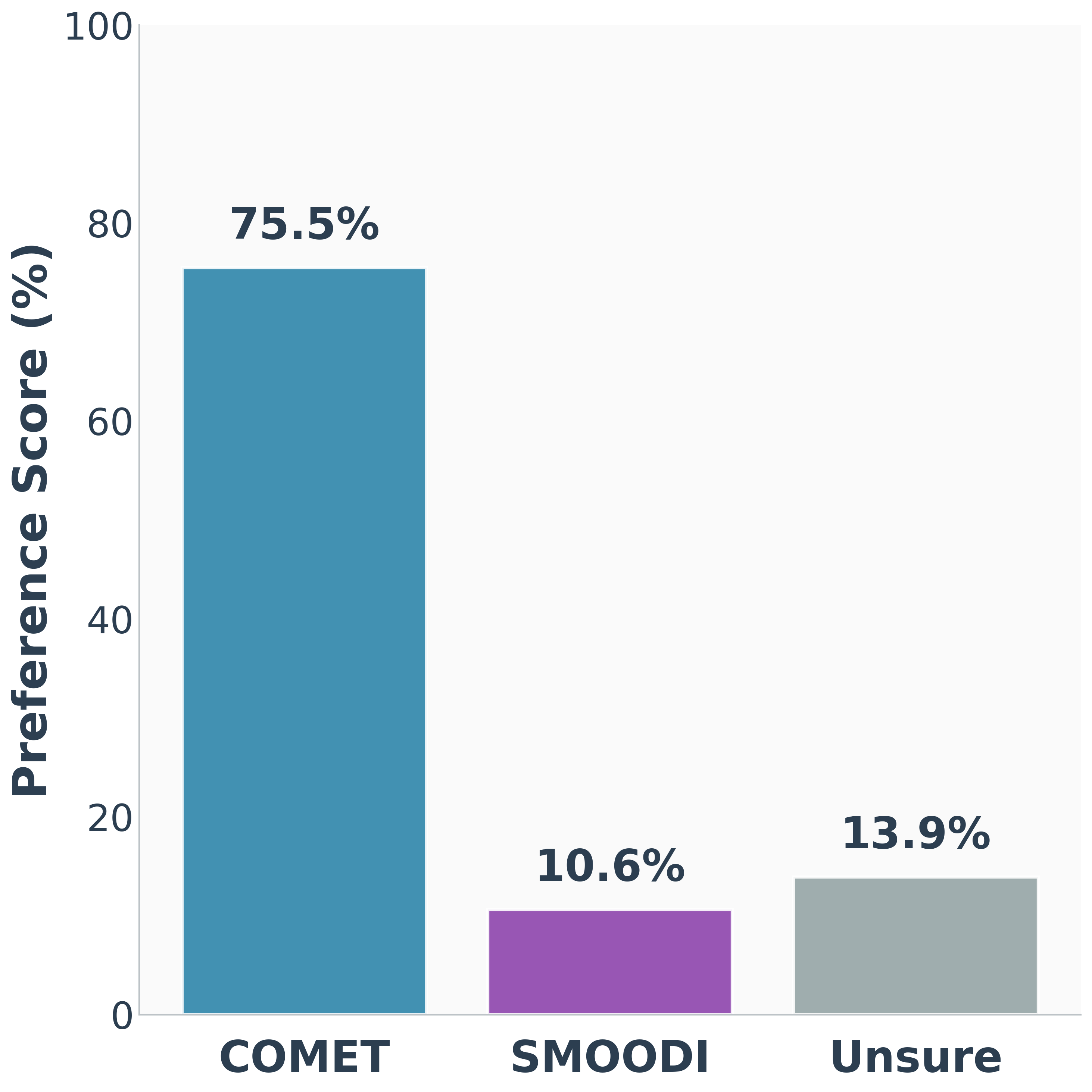}
    \caption{Motion Stylization}
    \label{fig:stylization_user_study}
\end{subfigure}
\caption{User preference study results for the Motion In-betweening and Motion Stylization tasks.}
\label{fig:style-user-preference}
\end{figure}

\paragraph{Motion In-betweening} Motion in-betweening is a fundamental task in character animation and motion synthesis, where the goal is to generate a smooth and realistic sequence of intermediate poses given a set of keyframes. Since COMET supports multi-joint control, it can be naturally extended from the multi-joint goal-reaching task to handle motion in-betweening. The key difference is that the starting pose is fixed to the given keyframe. Because in-betweening requires the generated motion to terminate precisely at the final target pose, we configure the Reference-Guided Feedback (RGF) to pull all controllable joints directly toward the final pose, ensuring seamless convergence. We evaluate performance following the protocol of Harvey et al.~\cite{harvey2020robust}. L2Q measures global quaternion error, L2P computes global positional error, and NPSS~\cite{gopalakrishnan2019neural} quantifies perceptual similarity via normalized power spectrum similarity. In addition to quantitative metrics, we conduct a user preference study (Figure~\ref{fig:mib_user_study}), where participants favor COMET over the baselines, confirming that its generated motions are perceptually more natural.

\paragraph{Motion Stylization}

To evaluate stylization quality, we conducted a user preference study. Participants were presented with side-by-side comparisons of motions generated by COMET and competing baselines, both stylized using the same reference sequences, and were asked to select the motion they found more natural and expressive, with a ``Unsure’’ option included. Figure~\ref{fig:stylization_user_study} summarizes the results. COMET was strongly preferred over competing method.

\begin{figure}[tbp]
  \centering 
  \begin{subfigure}[b]{0.8\linewidth} 
    \centering 
    \includegraphics[width=\linewidth]{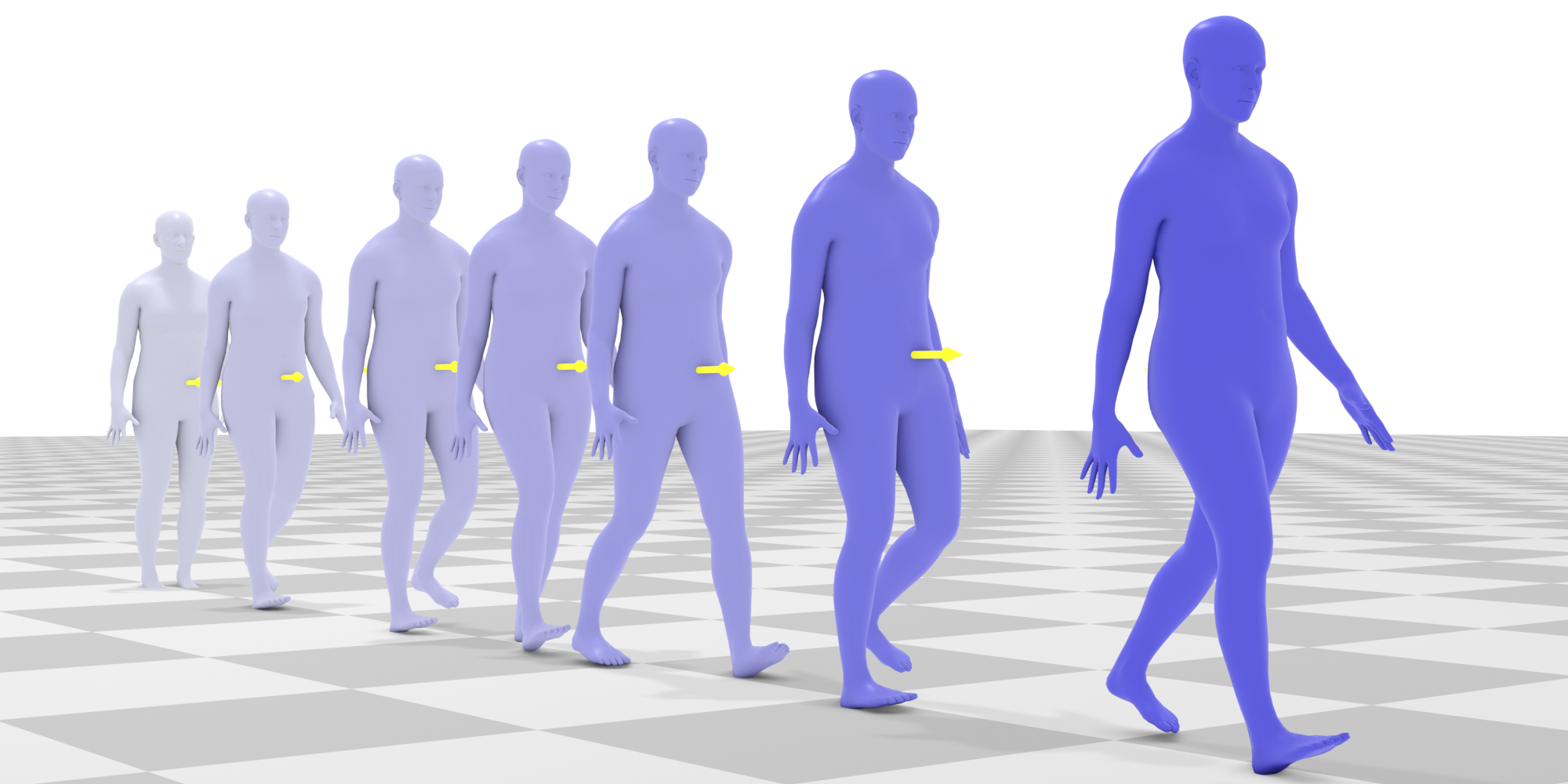}
    \caption{Content Motion (Walk)}
    \label{fig:style_walk_sub}
  \end{subfigure}
  \begin{subfigure}[b]{0.8\linewidth} 
    \centering 
    \includegraphics[width=\linewidth]{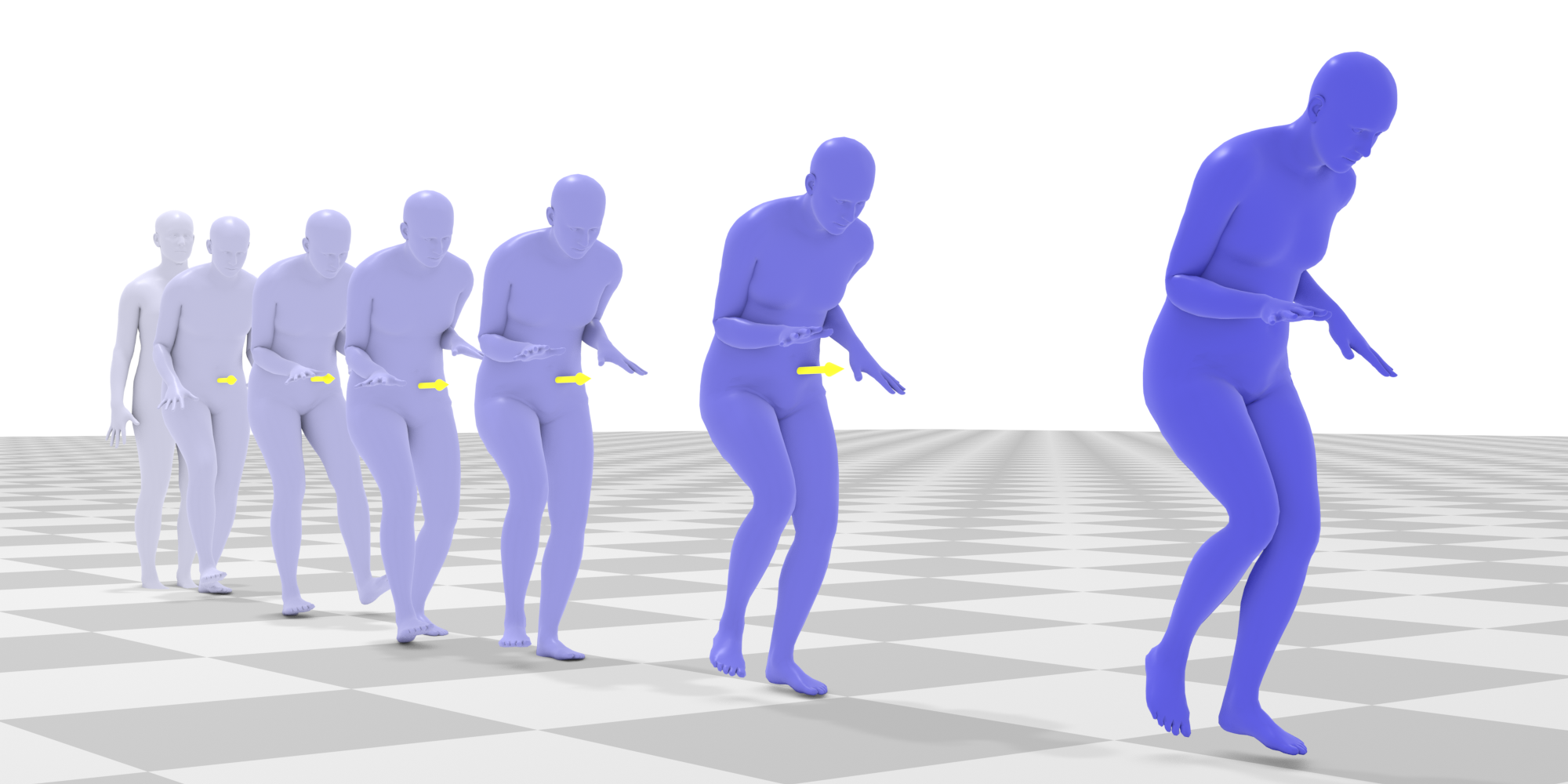}
    \caption{Stylization - Dinosaur}
    \label{fig:style_dinosaur_sub}
  \end{subfigure}
  \hfill 
  \begin{subfigure}[b]{0.8\linewidth} 
    \centering 
    \includegraphics[width=\linewidth]{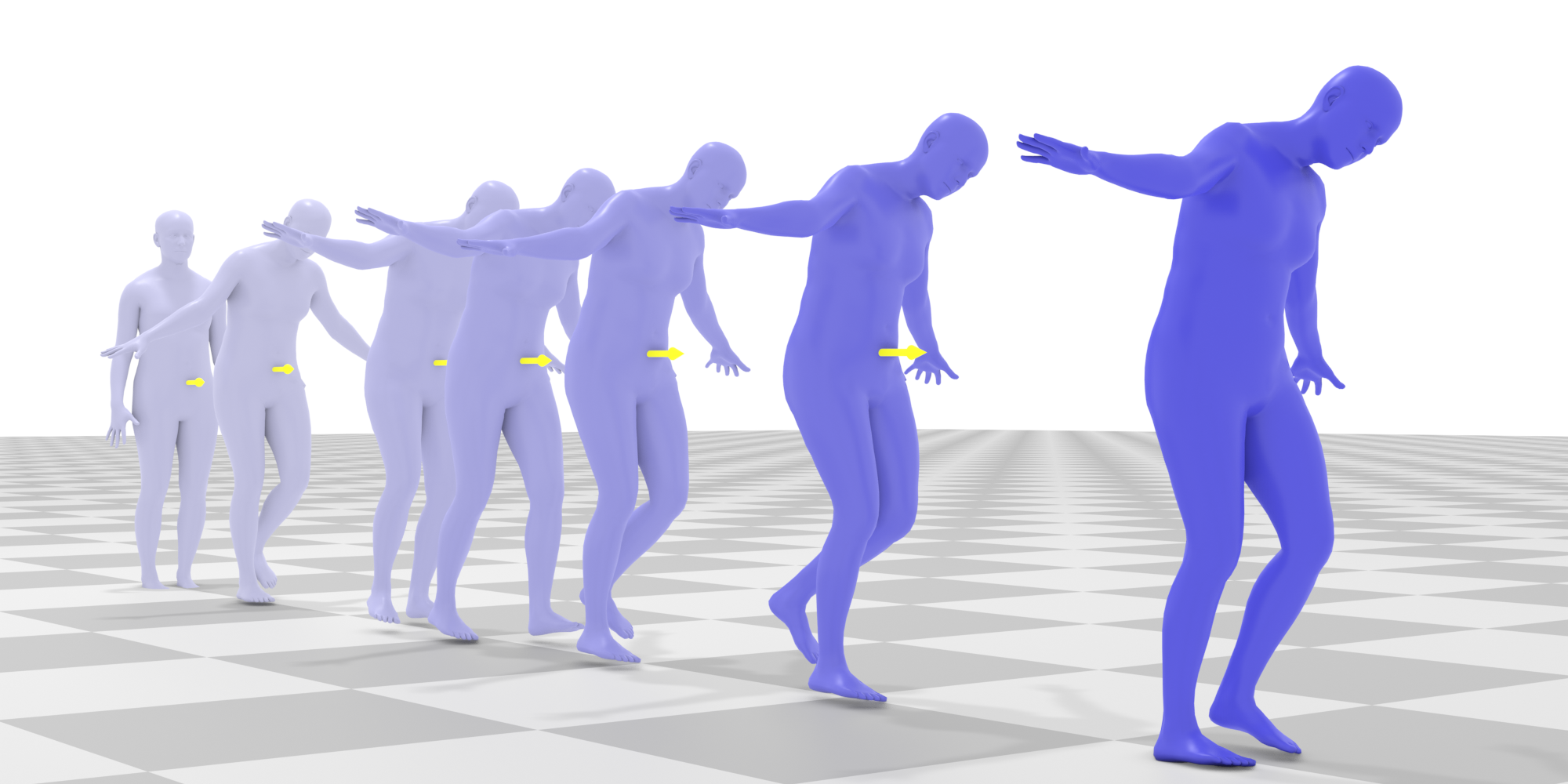}
    \caption{Stylization - Aeroplane}
    \label{fig:style_aeroplane}
  \end{subfigure}
  \caption{Motion Stylization via Reference-guided Feedback}
  \label{fig:stylization}
\end{figure}

%% file: sec/5_conclusion.tex
\section{Conclusion}
\label{sec:5_conclusion}

We introduce COMET, a real-time, autoregressive framework for controllable long-horizon human motion. COMET unifies fine-grained joint-level control and temporal robustness within a single conditional VAE–Transformer, enabled by two core components: (i) \emph{joint-wise attention} that conditions on arbitrary subsets of user-specified target joints, and (ii) a \emph{reference-guided feedback} loop that counteracts drift by gently attracting predictions toward a learned manifold of natural poses. This design yields a flexible model that supports single- and multi-joint goal reaching, motion in-betweening, and plug-and-play stylization without retraining. Extensive experiments demonstrate consistent gains in benchmark with ablations confirming the complementary roles of joint-wise attention for precise control and RGF for long-term stability.

\paragraph{Limitations and future work.}
While COMET is robust and efficient, there remain several promising directions for future work. Incorporating explicit contact and physics constraints could further reduce foot skate and enhance physical interactions with the environment. Extending COMET to process higher-level semantic control signals, such as task descriptions or natural language commands, would enhance its versatility and scalability.